%% file: acl_latex.tex
\documentclass[10pt]{article}
\usepackage{lineno}
\usepackage[preprint]{tmlr}

\usepackage{hyperref}
\usepackage{url}
\usepackage[utf8]{inputenc}
\usepackage{microtype}
\usepackage{inconsolata}
\usepackage{graphicx}
\usepackage{dsfont}      
\usepackage{xcolor}
\usepackage{xspace}
\usepackage{amsmath}
\usepackage{enumitem}
\usepackage{tcolorbox}
\usepackage{caption}
\usepackage{wrapfig}
\usepackage{booktabs}

\newcommand{\stitle}[1]{\vspace{2mm}\noindent{\bf #1.}}

\title{Synthetic Hallucinations, Real Gains: Hard Negatives from
       Frontier Models for FIM Hallucination Mitigation}

\author{\name Mahdi Erfanian \addr University of Illinois Chicago\thanks{Work done during a project at Microsoft CoreAI.} \email merfan2@uic.edu
      \AND
      \name Nelson Daniel Troncoso 
      \addr Microsoft
      \AND
      \name Aashna Garg 
      \addr Microsoft
      \AND
      \name Amabel Gale 
      \addr Microsoft
      \AND
      \name Xiaoyu Liu 
      \addr Microsoft
      \AND
      \name Pareesa Ameneh Golnari 
      \addr Microsoft
      \AND
      \name Shengyu Fu 
      \addr Microsoft
      }

\def\month{MM}
\def\year{YYYY}
\def\openreview{\url{https://openreview.net/forum?id=XXXX}}

\begin{document}
\maketitle

\input{sections/abstract}

\input{sections/introduction}

\input{sections/related_work}

\input{sections/method}

\input{sections/experiments}

\input{sections/analysis}

\input{sections/limitations}

\input{sections/conclusion}

\bibliography{custom}
\bibliographystyle{tmlr}

\input{sections/appendix}

\end{document}

%% file: sections/abstract.tex
\begin{abstract}
Small open-source code models that power IDE autocomplete still
emit \emph{hallucinated} Fill-in-the-Middle (FIM) completions:
syntactically natural calls to methods, parameters, variables, and
imports that do not exist in the surrounding project. Existing
mitigations either require per-language execution sandboxes that do
not apply at mid-keystroke or preference-optimisation pipelines that
need large human-labelled corpora. We propose an execution-free
alternative: use frontier code models to \emph{synthesise}
plausible-but-wrong completions as \emph{hard negatives}, then
leverage the contrast between these synthetic hallucinations and the
ground-truth developer edit as a supervised fine-tuning signal.
Our pipeline scrapes multilingual FIM contexts from public GitHub
across eight languages and asks a panel of three frontier generators
to produce one hard negative per context for each of four
hallucination types drawn from the Delulu taxonomy,
a Docker-verified multilingual FIM hallucination benchmark, yielding
a paired
\textsc{chosen}/\textsc{rejected} dataset. Fine-tuning
\textsc{Qwen2.5-Coder-7B-Instruct} on a 100K-row curated subset
lifts Delulu exact match by $+18.8$ points and edit similarity by
$+0.22$ on every language and every type, while also improving every
HumanEval-Infilling split and every SAFIM subset. The same recipe at
3B lifts Delulu by $+12.8$ EM with a small, characterised
general-FIM trade-off. Five-axis ablations (size, type mix, language
coverage, base-model family, and a difficulty-aware \emph{fool rate})
plus a head-to-head SFT vs.\ DPO/ORPO comparison map which design
choices drive the gain. We release the full pipeline source code ---
generation, fool-rate LLM judging, curation, and the FIM fine-tuning
recipe --- so that the experiments in this paper can be reproduced
end-to-end on any permissively licensed corpus.
\end{abstract}

%% file: sections/introduction.tex
\section{Introduction}
\label{sec:intro}

Large language models have become the default engine behind modern
code assistants, and one of the dominant interaction patterns is
\emph{Fill-in-the-Middle}
(FIM)~\citep{bavarian2022fim,fried2023incoder}: given a prefix and a
suffix from the same source file, the model produces the missing
middle. FIM is the workhorse of editor autocomplete: every
keystroke that opens a new gap between two pieces of code becomes,
at the model's input, a FIM request, and its quality directly
shapes the day-to-day experience of an AI coding assistant.

Because autocomplete must feel ``live'' (suggestions are useful only
if they arrive within a few hundred milliseconds of a
keystroke~\citep{ziegler2022productivity,mozannar2024realhumaneval}),
frontier APIs are too slow and too expensive at the scale of
millions of keystrokes per second, and increasingly unusable inside
enterprise tenants where source code cannot leave the customer's
network. Production code assistants therefore ship \emph{small,
fast, open-source code models} in the $3$--$7$B parameter range
specifically tuned for FIM~\citep{bavarian2022fim,fried2023incoder,guo2024deepseekcoder,yang2024qwen25coder,lozhkov2024starcoder2},
reserving larger frontier models for chat and explain workflows.

\begin{figure}[t]
  \centering
  \includegraphics[width=0.5\textwidth]{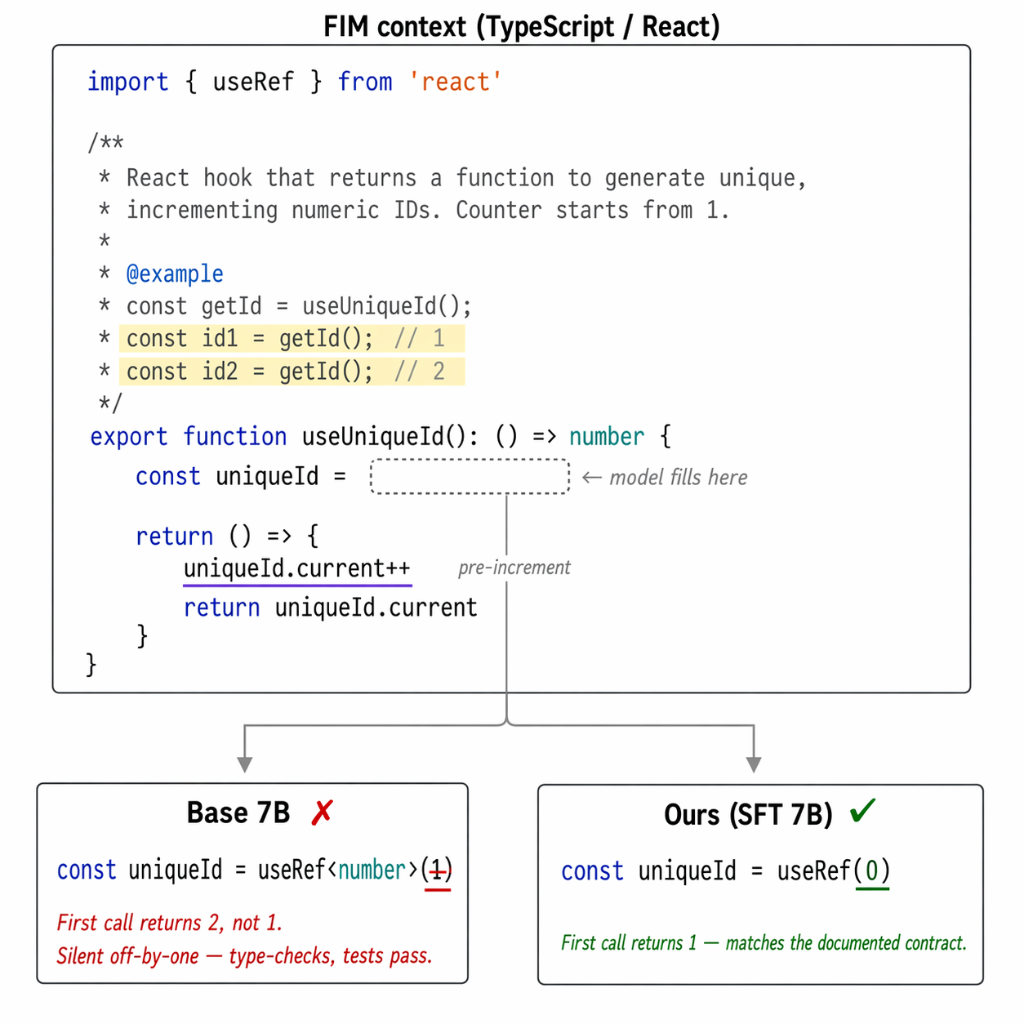}
  \caption{A representative FIM hallucination from Delulu. Given the
  same prefix and suffix (top), a React \texttt{useUniqueId} hook
  whose docstring promises that the first call returns \texttt{1} and
  whose body pre-increments the counter before returning it. The
  base 7B Code-LLM (bottom left) fills the hole with
  \texttt{useRef<number>(1)}, so the first call returns \texttt{2} and
  silently violates the documented contract. Our SFT-tuned model
  (bottom right) emits \texttt{useRef(0)}, matching the developer's
  actual edit and the docstring. The failure is local and plausible:
  the surrounding lines type-check and run, the hallucinated
  initialiser looks deliberate, and a unit test asserting only that
  IDs are increasing would not catch the off-by-one.}
  \label{fig:motivation}
\end{figure}

The persistent failure mode of such small FIM models is
\emph{code hallucination}: a completion that is syntactically natural and
locally plausible, but that violates the project's true state:
calling a method that does not exist, invoking an identifier that
was never imported, fabricating a positional argument the API does
not accept, or supplying an argument value that contradicts the
documented behaviour of the surrounding code
(Figure~\ref{fig:motivation}).

Two families of mitigations are visible in prior work. The first,
\emph{execution-based} curation and filtering, runs each candidate
completion through unit tests inside a per-language sandbox and
keeps only the ones that
pass~\citep{chen2021codex,liu2024execution,tian2024codehalu}; it is
the recipe behind HumanEval-style evaluation and behind
self-improving synthetic-data pipelines such as Magicoder and
WizardCoder~\citep{wei2024magicoder,luo2024wizardcoder}. The
approach is effective for full-function synthesis with tests
attached, but it breaks down in FIM autocomplete: mid-file edits
have no per-hole tests, the file often does not parse mid-keystroke,
and per-language sandboxes do not scale across multiple languages. The second family, \emph{preference optimisation}, trains
the model to prefer a chosen completion over a rejected one, either
through a learned reward model
(RLHF/RLAIF~\citep{lee2024rlaif}) or by collapsing the reward into a
contrastive objective (DPO~\citep{rafailov2023dpo},
ORPO~\citep{hong2024orpo}). These methods are powerful but require
large human-annotated preference corpora or per-rollout reward
signals, neither of which is available at the granularity that
per-hole hallucination correction needs.

A third option is suggested by work on \emph{hard-negative
mining}~\citep{schroff2015facenet,robinson2021hardnegatives,xiong2021ance}
and \emph{synthetic data} from strong
models~\citep{wang2023selfinstruct,wei2024magicoder,luo2024wizardcoder}.
Hard-negative mining trains a model by contrasting positive examples
with carefully chosen \emph{near-miss} negatives that are close
enough to be confusing but wrong in a specific, learnable way.
Meanwhile, synthetic-data distillation has shown that strong models
can transfer non-trivial capabilities to smaller students, including
instruction following, reasoning, and code generation. Most earlier code-focused synthetic-data pipelines still depend on
execution to verify their generations and therefore inherit the FIM
limitations above~\citep{wei2024magicoder,chen2023codex,liu2024execution}. We propose a different approach to overcome this
limitation: rather than generating \emph{correct} completions that
require an executor to validate, we invert the recipe and generate
\emph{wrong-but-plausible} completions as hard negatives. Every FIM
context already carries a \emph{golden} continuation---the
real line of code at the selected FIM hole---so we never need
an executor to decide which completion is correct. This shifts the
problem from ``synthesise a correct completion and verify it'' to
``synthesise a wrong-but-plausible completion as a hard negative to
contrast against the existing gold,'' removing the per-language
sandbox, the per-hole unit tests, and the file-must-parse
precondition in one move.

This raises the central question of this paper:

\begin{tcolorbox}[colback=gray!5, colframe=gray!60, boxrule=0.4pt, arc=1.5pt, left=4pt, right=4pt, top=3pt, bottom=3pt, boxsep=0pt]
\begin{center}
    \emph{Can synthetic hard negatives generated by frontier code models
serve as effective supervised fine-tuning signal for reducing
hallucinations in smaller open-source FIM models?}
\end{center}
\end{tcolorbox}

These synthetic hard negatives can be used in two ways. In
fine-tuning paradigms that consume paired
\textsc{chosen}/\textsc{rejected} samples such as
DPO~\citep{rafailov2023dpo} and ORPO~\citep{hong2024orpo}, the
hard negatives serve \emph{directly} as the rejected side of the
contrastive objective, pushing the model toward the golden
completion and away from the hallucinated one. SFT, by contrast,
only consumes the chosen continuation and cannot ingest a rejected
sample directly. Our claim is that the hard negatives are still useful
here \emph{indirectly}: because each one is a wrong answer that a
strong model believed was good enough to emit, the rate at which it
fools a blind LLM judge panel (our \emph{fool rate};
\S\ref{sec:experiments:ablations:foolrate}) is a difficulty signal
on the underlying FIM hole. Using this signal as a
hard-negative mining
filter~\citep{schroff2015facenet,robinson2021hardnegatives} lets us
train on a much smaller, harder slice of the chosen data without
losing quality.

Concretely, our pipeline samples FIM contexts from a permissive
multilingual corpus scraped from public GitHub across eight
languages (C\#, Go, Java, JavaScript, PHP, Python, Ruby, Rust). A panel of three frontier code
generators then produces one hard negative per context for
each of four identifier-level hallucination types:
\textsc{method}, \textsc{parameter}, \textsc{undefined-variable},
and \textsc{import}\footnote{We adopt these four categories from the
Delulu benchmark~\citep{delulu2026}, which finds them to cover the
majority of identifier-level FIM hallucinations. Extending the
taxonomy with finer-grained types is left to future work.}. The pipeline emits a
paired \textsc{chosen}/\textsc{rejected} dataset
(Table~\ref{tab:taxonomy}, \S\ref{sec:method}); we use the
\textsc{chosen} half as the SFT target and the \textsc{rejected}
half as the contrastive signal for the DPO/ORPO comparison reported
alongside SFT in the proof-of-concept evaluation
(\S\ref{sec:experiments:poc}).

Fine-tuning \textsc{Qwen2.5-Coder-7B-Instruct} on a 100K-row
curated slice of this data lifts Delulu exact match by $+18.8$
points ($+0.22$ edit similarity), with positive deltas in
\emph{every} language and \emph{every} hallucination type.
We evaluate on three complementary benchmarks beyond Delulu:
HumanEval-Infilling~\citep{fried2023incoder} (single-line,
multi-line, and random-span Python FIM),
SAFIM~\citep{gong2024safim} (API completion, block completion, and
control-flow completion across multiple languages), and
Real-FIM-Eval~\citep{gong2025realfimeval}, a benchmark of real
GitHub additions extracted from $228$ permissively-licensed
repositories.
The recipe simultaneously improves general-purpose FIM on every
HumanEval-Infilling split and every SAFIM subset. The same recipe applied to the smaller
\textsc{Qwen2.5-Coder-3B-Instruct} lifts Delulu by $+12.8$ EM with a
small, size-dependent trade-off on general FIM that we characterise
in Section~\ref{sec:analysis}. The recipe also internalises
\emph{stopping behaviour}: every curated training target ends
exactly at the FIM hole boundary, and a first-$N$-lines truncation
protocol that we apply to every table isolates this gain from
content gain (\S\ref{sec:experiments:setup}).

Having established that the recipe works, we ablate which design
choices drive it: training-set \emph{size} (5K--100K rows;
\S\ref{sec:abl:size}), the \emph{hallucination-type} mix
(\S\ref{sec:abl:type}), the \emph{language coverage} (Python-only,
top-5, all eight; \S\ref{sec:abl:lang}), the \emph{base-model
family} (Qwen, StarCoder2, CodeLlama; \S\ref{sec:abl:basemodels}),
and a difficulty-aware \emph{fool rate} that filters the training
set by hard-negative difficulty
(\S\ref{sec:experiments:ablations:foolrate}). The picture is
consistent: scale matters until $\sim$50K rows; all four
hallucination types are necessary at full scale; language breadth
ignites cross-lingual transfer above a five-language threshold,
the recipe is base-model-agnostic; and supervisory mass concentrates
in harder negatives (those that deceive more judges), with the
largest per-row gain coming from the disputed middle of the
fool-rate distribution rather than the unanimous-easy tier.

\paragraph{Contributions.}
\begin{itemize}[leftmargin=*]
  \item An \textbf{execution-free, multilingual hard-negative SFT
        pipeline} that turns a permissive code corpus and a frontier
        generator panel into paired
        $(\textsc{chosen}, \textsc{rejected})$ FIM data without
        sandboxes or unit tests (\S\ref{sec:method}).
  \item Open-source \textbf{training and judging code} ---
        the full pipeline used to generate hallucinated hard
        negatives, score them with a fast-reviewer LLM judge panel,
        curate the SFT dataset, and fine-tune the FIM models. The
        recipe is base-model-agnostic and we report results on four
        7B/3B base families
        (\textsc{Qwen2.5-Coder-\{3B,7B\}}, \textsc{StarCoder2-7B},
        \textsc{CodeLlama-7B}); the SFT recipe simultaneously reduces
        hallucinations and improves general FIM at 7B (with a small,
        characterised trade-off at 3B), and we report a head-to-head
        SFT vs.\ DPO/ORPO comparison on the same paired data
        (\S\ref{sec:experiments:poc}, \S\ref{sec:analysis:size},
        \S\ref{sec:analysis:sft},
        \S\ref{sec:abl:basemodels}).
  \item Evidence that the recipe also teaches \textbf{stopping
        behaviour}, isolated via a first-$N$-lines truncation scoring
        protocol applied to every table
        (\S\ref{sec:experiments:setup}).
  \item A systematic ablation of \textbf{five recipe axes}: size,
        hallucination-type mix, language coverage, base model, and
        fool-rate filtering (\S\ref{sec:experiments:ablations}).
\end{itemize}

\paragraph{Release.}
We open-source the full pipeline code --- generation prompts,
fool-rate LLM judge harness, curation, FIM tokenisation, and the
fine-tuning recipe --- so that any team with access to a permissively
licensed corpus and a frontier generator panel can reproduce the
proof of concept end to end. We do not release the fine-tuned
checkpoints or the curated dataset: both are built on a proprietary
source-code corpus whose redistribution terms prevent publication of
derivatives. For teams with limited compute, a reduced-scale recipe
using two generators and a $\sim\!50$K-row curated subset reaches
near-identical performance at roughly half the total cost
(Appendix~\ref{app:cost}). \S\ref{sec:method} and
Appendix~\ref{app:implementation} together document the pipeline,
the prompts, and the training hyperparameters in enough detail to
reproduce every result in this paper.

%% file: sections/related_work.tex
\section{Related Work}
\label{sec:related}

\paragraph{Code-LLMs and Fill-in-the-Middle.} Causal Code-LLMs such as
Codex~\citep{chen2021codex}, StarCoder2~\citep{lozhkov2024starcoder2},
DeepSeek-Coder~\citep{guo2024deepseekcoder}, and
Qwen2.5-Coder~\citep{yang2024qwen25coder} are routinely fine-tuned with
FIM token permutations~\citep{bavarian2022fim}. We adopt the same FIM
input format and evaluate on standard FIM benchmarks
(HumanEval-Infilling~\citep{fried2023incoder},
SAFIM~\citep{gong2024safim}).

\paragraph{Code hallucination benchmarks.}
CodeHalu~\citep{tian2024codehalu} and HalluCode~\citep{liu2024hallucode}
target instruction-following code tasks and rely on execution-based
verification to label correctness; neither provides a FIM-specific
evaluation suite, so no direct numerical comparison with our work
is possible.
Delulu~\citep{delulu2026}
provides a Docker-verified multilingual FIM benchmark with four
hallucination types and is the held-out evaluation target of our work.
We reuse Delulu's taxonomy but \emph{drop} Docker verification when
constructing training data, replacing the execution oracle with the
existence of a known-confusable hard negative produced by a
strong frontier generator.

\paragraph{Synthetic SFT data for code.}
Self-Instruct~\citep{wang2023selfinstruct},
Magicoder~\citep{wei2024magicoder}, and
WizardCoder~\citep{luo2024wizardcoder} construct large code SFT corpora
from LLM completions. The closest line of work is RLAIF-style
\citep{lee2024rlaif} pipelines that pair correct and incorrect
completions for DPO~\citep{rafailov2023dpo}. Unlike previous code-SFT
work, our \emph{target} failure mode is hallucination at FIM holes
specifically, and our paired data is built around a synthetic
hard negative produced by a strong generator rather than a synthetic
preference signal.

\paragraph{Test-free correctness signals.} Execution-based filters
\citep{chen2023codex,liu2024execution} rely on a per-language runtime.
At SFT scale (millions of rows, nine languages, multi-runtime tests)
this is prohibitive. Static substitutes (parsers, type-checkers) are
cheaper but miss semantic hallucinations. Our pipeline sidesteps both:
the \emph{hard negative} produced by a strong generator
provides the supervisory signal --- training the model to prefer the
golden over the hallucinated counterpart directly targets the failure
mode the hard negative represents.

\paragraph{LLM-as-a-judge.} Using LLMs as evaluators is now standard
practice across NLG~\citep{zheng2023judging,gu2024llmaaj}. Recent work
applies panels of judges to reduce single-model bias
\citep{verga2024panelofjudges}. We adopt judges only as a
\emph{difficulty signal} in an ablation
(\S\ref{sec:experiments:ablations:foolrate}); the main proof of concept
does not depend on judging.

%% file: sections/method.tex
\section{Method}
\label{sec:method}

\begin{figure}[t]
  \centering
  \includegraphics[width=\textwidth]{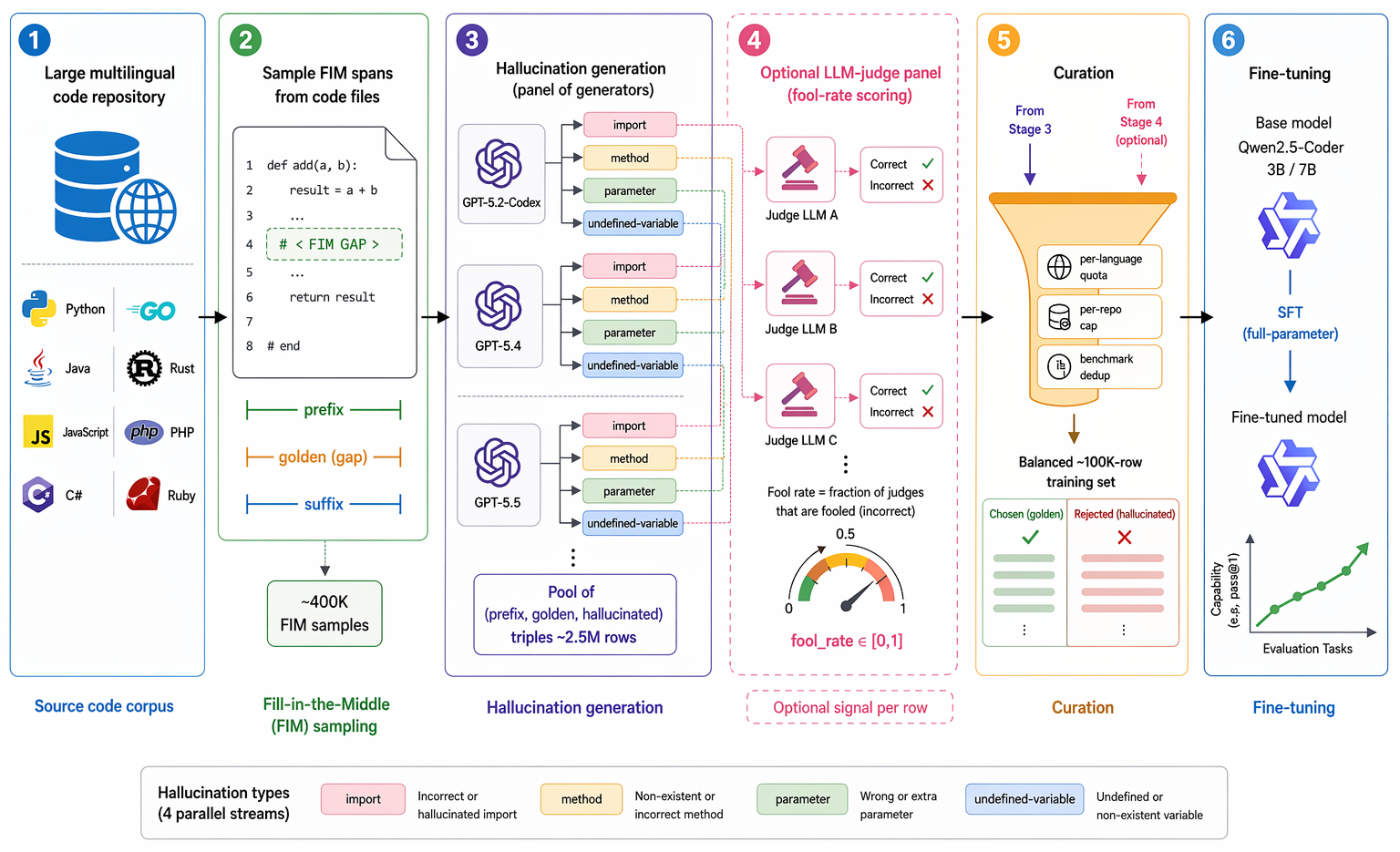}
  \caption{End-to-end pipeline. \textbf{(1)}~A multilingual source-code
  corpus spanning eight languages feeds \textbf{(2)}~Fill-in-the-Middle
  sampling, which extracts ${\sim}400$K real $(\text{prefix},
  \text{golden}, \text{suffix})$ contexts. \textbf{(3)}~A panel of
  three strong code generators (GPT-5.2-Codex, GPT-5.4, GPT-5.5) emits
  one hallucinated completion per context for each of the four
  taxonomy types, producing a pool of ${\sim}2.5$M
  $(\text{prefix}, \text{golden}, \text{hallucinated})$ triples.
  \textbf{(4)}~An \emph{optional} panel of blind LLM judges scores each
  row with a fool rate
  $\in [0,1]$ (\S\ref{sec:method:fool-rate}); the main proof of
  concept does not use this signal. \textbf{(5)}~Curation enforces
  per-language and per-repository quotas and emits a balanced
  ${\sim}100$K-row training set with both the golden completion
  (\textsc{chosen}) and the hallucinated one (\textsc{rejected}).
  \textbf{(6)}~Full-parameter SFT of Qwen2.5-Coder-\{3B,7B\} on the
  \textsc{chosen} side yields the fine-tuned models we evaluate.
  Engineering details are in Appendix~\ref{app:implementation}.}
  \label{fig:pipeline}
\end{figure}

Our pipeline (Figure~\ref{fig:pipeline}) turns a multilingual code
corpus into a paired \textsc{chosen}/\textsc{rejected} SFT dataset
without ever executing the candidate code. The four hallucination
types we target are listed in Table~\ref{tab:taxonomy}; they are
inherited from the Delulu benchmark~\citep{delulu2026} so that gains
on Delulu are interpretable as targeted-error reductions, not
distribution-shift artifacts.

\begin{table}[t]
  \centering\small
  \setlength{\tabcolsep}{4pt}
  \renewcommand{\arraystretch}{1.05}
  \begin{tabular}{@{}p{0.19\textwidth}p{0.79\textwidth}@{}}
  \toprule
  \textbf{Type} & \textbf{Edit applied to the golden completion}\\
  \midrule
  \textsc{method} &
    Replace one method name with an invented but plausible name not
    present in prefix or suffix.\\[2pt]
  \textsc{parameter} &
    Inject a fake keyword or positional argument into an existing call.\\[2pt]
  \textsc{undefined variable} &
    Replace one identifier with one not defined in the surrounding
    scope.\\[2pt]
  \textsc{import} &
    Replace one real import line with a plausible but fictitious
    package or symbol.\\
  \bottomrule
  \end{tabular}
  \caption{Hallucination taxonomy used throughout the paper.}
  \label{tab:taxonomy}
\end{table}

\subsection{Pipeline}
\label{sec:method:pipeline}

\paragraph{Seed FIM contexts.} We scrape permissively licensed public
GitHub repositories across eight languages and apply a model-based
selector to identify lines that make high-quality FIM completion holes
(sufficient context in prefix and suffix, non-trivial golden content).
The selected line becomes the golden completion; the surrounding code
forms the prefix and suffix. We sample $\sim$400K such contexts
stratified per language and exclude any file or repository present in
the evaluation benchmarks (\S\ref{sec:experiments:setup}).

\paragraph{Hallucination generation.} A panel of three strong
generators (GPT-5.2-Codex, GPT-5.4, GPT-5.5) produces one hallucinated
completion per $(\text{context}, \text{type})$ pair. Different
generators exhibit visibly different failure modes (one over-uses
camel-case verbs, another invents framework-specific decorators),
so panelling diversifies the hallucinations the model must learn to
suppress; we also use the panel as an ablation axis in
\S\ref{sec:experiments:ablations}. The \textsc{import} type is
special because the candidate must be placed at file scope rather
than at the FIM hole; we handle it by relocating the FIM hole onto
the original import line
(Appendix~\ref{app:implementation:import-restructure}).

\paragraph{Curation.} The curator emits two parallel files: an
\texttt{sft.jsonl} that pairs each FIM prompt with its golden
completion (used by the main proof of concept and by every ablation
in \S\ref{sec:experiments:ablations}), and a \texttt{pairs.jsonl}
that additionally retains the hallucinated completion as
\textsc{rejected}. Curation enforces per-(language,
type) bucket quotas (default $4{,}000$ rows) and a per-source-
repository cap (default $50$) to control leakage of any single repo's
style.

\paragraph{Fine-tuning.} We fine-tune
\textsc{Qwen2.5-Coder-\{3B, 7B\}-Instruct}~\citep{yang2024qwen25coder}
with full-parameter SFT on the curated \texttt{sft.jsonl}; the model
consumes the FIM prefix and suffix and predicts the golden completion.
The same recipe is applied to two additional 7B base families
(\textsc{StarCoder2-7B}, \textsc{CodeLlama-7b-hf}) in the
base-model ablation~(\S\ref{sec:abl:basemodels}); the only
per-family change is the FIM sentinel template.
Hyperparameters in Appendix~\ref{app:implementation:training}.

\subsection{Difficulty signal: fool rate}
\label{sec:method:fool-rate}

The pipeline above treats every generated hallucination as equally
useful training signal. In practice, some hallucinations are easy to
spot (an obviously fake method name like \texttt{doFakeStuff}) while
others are subtle enough that even a strong LLM judge would mistake
them for correct code. Intuitively, hard-to-detect hallucinations
should be more valuable for training: they are closer to the mistakes
the target model would actually make in the wild. To measure how
hard each row is, we score it with a panel of LLM judges and define
the \emph{fool rate} as the fraction of judges that classified the
hallucinated completion as correct. A fool rate of 1 means every
judge was deceived; a fool rate of 0 means the hallucination was
obvious to all. This score lets us select training rows by difficulty
and test empirically which regime carries the most useful signal.

Formally, we ask a panel
$\mathcal{J} = \{j_1, j_2, j_3\}$ of three fast-reviewer LLM judges
(\textsc{GPT-4o-mini}, \textsc{GPT-4.1-mini}, \textsc{GPT-5.4-mini};
chosen to be distinct from the three Phase-2 generators), each
shown only the FIM prefix, the candidate completion, and the suffix
\emph{without being told the candidate may be hallucinated},
whether the candidate is correct. Each verdict is a binary
correct/incorrect label and the \emph{fool rate} is the fraction of
judges deceived:
\begin{equation}
\label{eq:foolrate}
\mathrm{fool\_rate} \;=\; \frac{1}{|\mathcal{J}|} \sum_{j\in\mathcal{J}}
\mathds{1}\!\left[j(\text{candidate}) = \text{correct}\right].
\end{equation}
$\mathrm{fool\_rate}{=}1$ marks rows that every judge thought were
correct (the hardest); $\mathrm{fool\_rate}{=}0$ marks rows whose
hallucination is obvious. The fool rate is stored as a per-row column
and consumed by the difficulty ablation
(\S\ref{sec:experiments:ablations:foolrate}), which trains on
subsets thresholded by $\tau$ to test which intuition wins; the main
proof of concept (\S\ref{sec:experiments:poc}) does not filter on it.

%% file: sections/experiments.tex
\section{Experiments}
\label{sec:experiments}

Our experiments are organised around two questions.
\emph{Does the recipe work?} \S\ref{sec:experiments:poc} establishes
the proof of concept on Delulu and eight further FIM benchmarks at
two model scales (Qwen2.5-Coder-3B and 7B). \emph{Why does it work?}
\S\ref{sec:experiments:ablations} dissects the recipe along five
axes that map to the curator's free choices: training-set
\emph{size}~(\S\ref{sec:abl:size}), \emph{hallucination
type}~(\S\ref{sec:abl:type}), \emph{language
coverage}~(\S\ref{sec:abl:lang}), \emph{base-model
family}~(\S\ref{sec:abl:basemodels}), and a difficulty-aware
\emph{fool-rate threshold}~(\S\ref{sec:experiments:ablations:foolrate}).
A short 3B re-run (\S\ref{sec:abl:size3b}) confirms that the
size/type/language conclusions are not specific to the 7B scale.
\S\ref{sec:experiments:setup} describes the shared setup used
throughout.

\subsection{Setup}
\label{sec:experiments:setup}

\begin{figure}[t]
  \begin{minipage}[t]{0.48\textwidth}
  \vspace{0pt}
  \centering\footnotesize
  \begin{tabular}{lrr}
  \toprule
  \textbf{Language} & \textbf{Seed rows} & \textbf{Curated rows} \\
  \midrule
  C\#         & 71{,}832 & 15{,}938 \\
  Java        & 55{,}122 & 14{,}793 \\
  PHP         & 53{,}897 & 14{,}756 \\
  Go          & 51{,}008 & 14{,}832 \\
  JavaScript  & 49{,}710 & 14{,}079 \\
  Python      & 47{,}241 & 12{,}353 \\
  Ruby        & 35{,}406 & \phantom{0}7{,}117 \\
  Rust        & 35{,}340 & \phantom{0}6{,}018 \\
  \midrule
  \textbf{Total} & \textbf{399{,}556} & \textbf{99{,}886} \\
  \bottomrule
  \end{tabular}
  \captionof{table}{Per-language row counts: Phase~1 seed FIM contexts
  (left) and Phase~5 curated training rows (right). TypeScript is
  omitted throughout: the source corpus contains only $\sim$300
  TypeScript seeds, too small to affect curated proportions.}
  \label{tab:corpus-langs}
  \end{minipage}\hfill
  \begin{minipage}[t]{0.48\textwidth}
  \vspace{0pt}
  \centering
  \includegraphics[width=\textwidth]{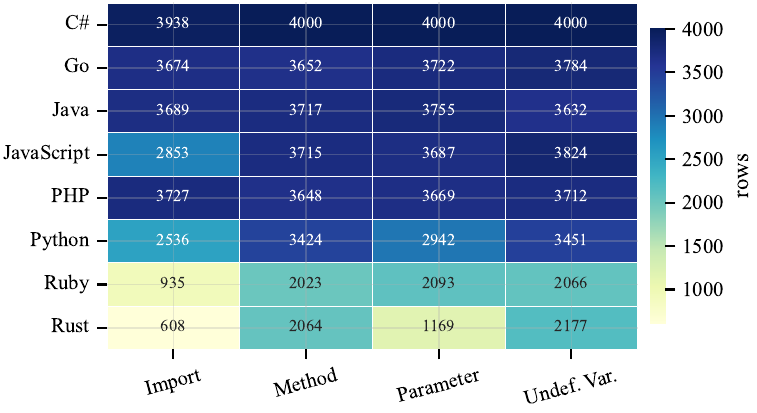}
  \captionof{figure}{Curated training set distribution per (language,
  hallucination type). C\# and Java are the largest cells; sampling
  is close to uniform across the four types within each language.}
  \label{fig:curated-heatmap}
  \end{minipage}
\end{figure}

\stitle{Source FIM corpus} We sample
$N_{\mathrm{seed}}=399{,}556$ FIM contexts from a permissive,
multilingual code corpus that already provides a fixed
$(\mathrm{prefix}, \mathrm{golden}, \mathrm{suffix})$ (i.e.\ we do not synthesise FIM
holes; we sample existing ones, consistent with our research
question). The corpus covers eight programming languages
(Table~\ref{tab:corpus-langs}, ``seed'' column); rows are stratified
per language and we exclude any file or repository present in the
Delulu held-out benchmark to avoid evaluation contamination.

\stitle{Generation} For every seed context, each generator in
the panel $\{\text{GPT-5.2-Codex}, \text{GPT-5.4}, \text{GPT-5.5}\}$
produces one hallucinated completion for each of the four taxonomy
types. After discarding rows that failed the output contract or the
import-type file-reconstruction check
(Appendix~\ref{app:implementation:import-restructure}), the resulting
pool contains $2{,}473{,}312$ valid (golden, hallucinated, context)
rows (Table~\ref{tab:gen_volume}, Appendix~\ref{app:data}). This is
the shared pool from which every ablation cell in
\S\ref{sec:experiments:ablations} samples its training rows.

\stitle{Curation} Phase~5 distils the generation pool into a
balanced $100{,}000$-row training set under the following
constraints:
\begin{itemize}[leftmargin=*]\itemsep0pt
  \item Per-(language, type) bucket cap of $4{,}000$ (so any cell
        contributes at most $\sim$4\% of the dataset).
  \item Per-source-repository cap of $50$ rows.
  \item Delulu-overlap exclusion on \texttt{file\_path} and
        \texttt{source\_repo}.
\end{itemize}
The curated dataset is multilingual (Table~\ref{tab:corpus-langs}).
We emit both \texttt{sft.jsonl} (golden as supervised target) and
\texttt{pairs.jsonl} (golden + hallucinated) for downstream
preference-optimisation experiments; the experiments in this section
use \texttt{sft.jsonl}. Per-language curated counts are the right
column of Table~\ref{tab:corpus-langs}.

\stitle{Training} We fine-tune
\textsc{Qwen2.5-Coder-\{3B, 7B\}-Instruct}~\citep{yang2024qwen25coder}
with full-parameter fine-tuning. Per-device batch size $4$, gradient
accumulation $8$, effective batch $32$ on a single $8\times$H100 node,
learning rate $5\times 10^{-6}$, cosine schedule, $1$ epoch,
$\mathrm{cutoff\_len}=8000$, DeepSpeed ZeRO-3, bf16, FlashAttention-2.
Full hyperparameters in Appendix~\ref{app:implementation:training}.

\stitle{Metrics} We report two metrics throughout. \textbf{EM}
(exact match, \%) is the fraction of predictions that match the
golden completion exactly after whitespace normalisation; it is the
strictest signal and the one we treat as primary. \textbf{ES} (edit
similarity, $[0,1]$) is the character-level ratio returned by
Python's \texttt{difflib.SequenceMatcher} against the golden
completion; it captures partial improvements that EM misses (near-
misses, formatting variants) and is more stable on small per-cell
counts. We use EM as the headline number and ES as the secondary
view; the two together also let us read off when a recipe shifts
content vs.\ surface form (e.g.\ the fool-rate ablation,
\S\ref{sec:experiments:ablations:foolrate}).

\stitle{Evaluation benchmarks} We evaluate on four benchmark
families (nine splits in total, $\sim$42K examples). \textbf{Delulu}
is our primary hallucination metric and is multilingual by design;
the remaining three families probe whether the recipe damages general
FIM ability on data that has nothing to do with hallucinations.
\begin{itemize}\itemsep0pt
  \item \textbf{Delulu}~\citep{delulu2026} ($N{=}1{,}950$, 7
        languages, 4 hallucination types) --- pairs every prompt with
        a hallucinated hard negative; EM and ES against the
        golden completion measure how often the model resists the
        hard negative.
  \item \textbf{HumanEval-Infilling}~\citep{fried2023incoder}
        ($N{=}409$, three splits: single-line, multi-line,
        random-span) --- canonical short-Python FIM regression test.
  \item \textbf{SAFIM}~\citep{gong2024safim} ($N{=}22{,}291$, four
        subsets: API call, code block, code block v2, control flow)
        --- a broader, enterprise-style FIM benchmark.
  \item \textbf{Real-FIM-Eval (add)}~\citep{gong2025realfimeval}
        ($N{=}17{,}879$): real code additions extracted from GitHub
        commits across $228$ permissively licensed repositories
        (Jan--Feb 2025). The ``edit'' split uses a conflict-merge
        format outside the standard FIM paradigm and is omitted.
\end{itemize}

\stitle{Inference} All evaluations use vLLM, greedy decoding
($T{=}0$, $\mathrm{max\_new\_tokens}=256$). We never use the Phase-2 generators (the frontier models that
produced the training data) at evaluation time, and never train on
any sample from any evaluation benchmark.

\stitle{Compute} The full project (proof of concept, all five
ablations, and discarded preliminary runs) consumed $\sim$$1{,}100$
H100-hours over $554$ AML jobs: $\sim$$640$ on training
($8{\times}$H100 nodes) and $\sim$$500$ on inference
($8{\times}$H100 with vLLM). Per-(model, training-set) SFT runs
averaged about an hour wall-clock and per-(model, benchmark) eval
cells $\sim$$8$ minutes. A bottom-up breakdown by job kind, base
model, and reduced-scale recipe is in Appendix~\ref{app:cost}.

\stitle{First-$N$-lines truncation} EM is a brittle metric for FIM
because a base model that produces the correct first line followed by
extra tokens scores $0$, even when its top-line completion is exactly
right. To separate this ``content vs.\ stopping'' confound from the
true content lift, we report every result table both
\emph{untruncated} and with first-$N$-lines truncation
(``+trunc''): given gold completion $g$, we keep only the first
$\mathrm{lines}(g)$ lines of the prediction before applying EM. This
upper-bounds what a perfect length oracle would buy a base model.
Truncation is computed at scoring time only --- decoding is unchanged
--- and uses the line count (not character count) so that whitespace
changes upstream do not break it. Critically, our SFT recipe trains
the model to emit the correct \emph{ending}, not just the correct
content: every curated row's target completion ends exactly where the
golden FIM hole ends, and the supervised objective penalises over- or
under-emission. As a result, the gap between the untruncated and
+trunc rows of every SFT cell in
\S\S\ref{sec:experiments:poc}--\ref{sec:abl:basemodels} is
$\leq\!1$~EM, while the same gap on base models is regularly
$+5$--$+25$~EM. The +trunc rows therefore serve a dual purpose:
they let us compare against the base on equal stopping footing, and
they quantify the stopping signal that SFT internalises.

\subsection{Proof of concept}
\label{sec:experiments:poc}

\begin{figure}[t]
  \begin{minipage}[t]{0.48\textwidth}
  \vspace{0pt}
  \centering\small
  \setlength{\tabcolsep}{4pt}
  \begin{tabular}{lcc}
  \toprule
  \textbf{Model} & \textbf{EM} & \textbf{ES} \\
  \midrule
  Qwen2.5-Coder-3B (base)         & $45.7{\pm}0.2$ & $0.72{\pm}0.02$ \\
  \quad +trunc                    & $52.0$         & $0.798$ \\
  \quad + SFT (ours)              & $\mathbf{58.5{\pm}0.3}$ & $\mathbf{0.85{\pm}0.02}$ \\
  \quad +trunc                    & $\mathbf{59.2}$ & $\mathbf{0.835}$ \\
  \midrule
  Qwen2.5-Coder-7B (base)         & $42.8{\pm}0.1$ & $0.65{\pm}0.02$ \\
  \quad +trunc                    & $48.8$         & $0.770$ \\
  \quad + SFT (ours)              & $\mathbf{61.4{\pm}0.3}$ & $\mathbf{0.87{\pm}0.02}$ \\
  \quad +trunc                    & $\mathbf{61.8}$ & $\mathbf{0.850}$ \\
  \bottomrule
  \end{tabular}
  \captionof{table}{Delulu held-out ($N{=}1{,}950$, 7 languages, 4
  hallucination types): EM (\%) and ES.
  Values are mean$\pm$half-spread across two inference runs;
  ``+trunc'' applies first-$N$-lines truncation
  (\S\ref{sec:experiments:setup}). The trunc lifts on base
  ($+6.3$~/~$+6.0$~EM at 3B/7B) vanish after SFT
  ($+0.7$~/~$+0.4$): SFT teaches the model to stop on its own.}
  \label{tab:delulu-headline}
  \end{minipage}\hfill
  \begin{minipage}[t]{0.48\textwidth}
  \vspace{0pt}
  \centering
  \includegraphics[width=\textwidth]{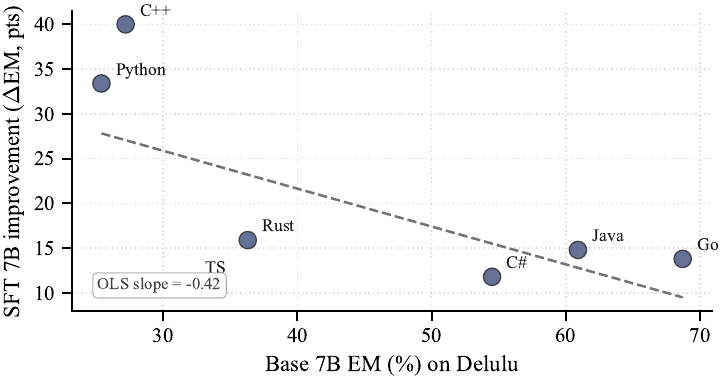}
  \captionof{figure}{Per-language Delulu gain ($\Delta$EM) for the 7B
  SFT model vs.\ the base 7B's baseline strength on the same
  language. The fitted OLS line has a strong negative slope: the
  recipe closes the largest gaps first.}
  \label{fig:delulu-gain-vs-base}
  \end{minipage}
\end{figure}

We fine-tune the recipe on Qwen2.5-Coder-3B and 7B and evaluate
without further tuning on Delulu and eight further FIM benchmark
splits. Three views answer three questions:
\emph{does it move the headline metric?}
(Table~\ref{tab:delulu-headline}),
\emph{does the gain distribute across hallucination types and
languages, or concentrate in one corner?}
(Table~\ref{tab:delulu-breakdown},
Figure~\ref{fig:delulu-gain-vs-base}), and
\emph{does the recipe transfer to FIM benchmarks it was not designed
for, and how does it compare to contrastive recipes on the same paired
data?} (Table~\ref{tab:cross-benchmark}).

\stitle{Headline} Both fine-tuned models improve substantially on
Delulu (Table~\ref{tab:delulu-headline}). Notably, \textsc{3B + SFT}
($58.5$ EM) outperforms \textsc{base 7B} ($42.8$ EM) by $15.7$ points:
the training data, not the parameter count, is doing the work.

\begin{table}[t]
  \centering\footnotesize
  \setlength{\tabcolsep}{3pt}
  \begin{minipage}[t]{0.48\textwidth}
  \centering
  \textit{By hallucination type}\vspace{2pt}\\
  \begin{tabular}{lcc|cc}
  \toprule
   & \multicolumn{2}{c|}{\textbf{3B}} & \multicolumn{2}{c}{\textbf{7B}}\\
                                & base & +SFT & base & +SFT \\
  \midrule
  \textsc{import}              & 40.7 & \textbf{62.5} & 39.4 & \textbf{65.6} \\
  \quad +trunc                 & 52.0 & \textbf{62.7} & 48.8 & \textbf{65.6} \\
  \textsc{method}              & 44.0 & \textbf{54.4} & 41.4 & \textbf{58.1} \\
  \quad +trunc                 & 48.6 & \textbf{55.3} & 46.4 & \textbf{58.4} \\
  \textsc{parameter}           & 51.7 & \textbf{62.1} & 48.3 & \textbf{65.1} \\
  \quad +trunc                 & 56.1 & \textbf{62.5} & 52.6 & \textbf{65.1} \\
  \textsc{undef-variable}      & 47.3 & \textbf{56.5} & 43.0 & \textbf{58.9} \\
  \quad +trunc                 & 51.6 & \textbf{57.0} & 47.7 & \textbf{58.9} \\
  \bottomrule
  \end{tabular}
  \end{minipage}\hfill
  \begin{minipage}[t]{0.48\textwidth}
  \centering
  \textit{By target language}\vspace{2pt}\\
  \begin{tabular}{lcc|cc}
  \toprule
   & \multicolumn{2}{c|}{\textbf{3B}} & \multicolumn{2}{c}{\textbf{7B}}\\
                                & base & +SFT & base & +SFT \\
  \midrule
  cpp           & 36.8 & \textbf{60.0} & 27.2 & \textbf{67.2} \\
  \quad +trunc  & 46.4 & \textbf{60.8} & 32.0 & \textbf{67.2} \\
  csharp        & 56.1 & \textbf{63.8} & 54.5 & \textbf{66.3} \\
  \quad +trunc  & 58.5 & \textbf{63.8} & 58.5 & \textbf{66.3} \\
  go            & 73.9 & \textbf{80.8} & 68.7 & \textbf{82.5} \\
  \quad +trunc  & 76.3 & \textbf{80.8} & 70.1 & \textbf{82.5} \\
  java          & 61.7 & \textbf{72.8} & 60.9 & \textbf{75.7} \\
  \quad +trunc  & 69.1 & \textbf{72.8} & 69.1 & \textbf{75.7} \\
  python        & 31.8 & \textbf{56.7} & 25.4 & \textbf{58.8} \\
  \quad +trunc  & 44.9 & \textbf{56.7} & 38.5 & \textbf{58.3} \\
  rust          & 38.2 & \textbf{50.2} & 36.3 & \textbf{52.2} \\
  \quad +trunc  & 42.6 & \textbf{53.8} & 43.8 & \textbf{53.4} \\
  typescript    & 31.2 & \textbf{38.8} & 32.1 & \textbf{43.3} \\
  \quad +trunc  & 35.0 & \textbf{38.8} & 33.6 & \textbf{43.3} \\
  \bottomrule
  \end{tabular}
  \end{minipage}
  \caption{Delulu EM (\%) per hallucination type (left) and per target
  language (right). Every cell improves; the largest absolute gains
  land where the base was weakest (\textsc{import}; C++, Python). C++
  and TypeScript have zero curated training rows
  (Table~\ref{tab:corpus-langs}) and still improve through
  cross-lingual transfer.}
  \label{tab:delulu-breakdown}
\end{table}

\stitle{The gain is broad, not local}
Table~\ref{tab:delulu-breakdown} breaks the Delulu gain down by
hallucination type (top) and target language (bottom). All four types
and all seven languages improve at both scales; the largest absolute
gains accrue where the base model was weakest: \textsc{import} at the
type level ($+22$ / $+26$ EM at 3B / 7B) and C++ and Python at the
language level ($+23$ to $+40$ EM, with the largest lift on C++ at
7B and on Python at 7B).
Figure~\ref{fig:delulu-gain-vs-base} makes this gap-closing pattern
explicit: regressing per-language gain on base-model strength gives a
strong negative slope.

\stitle{Cross-benchmark transfer, and SFT vs.\ DPO vs.\ ORPO on the
same paired data} Table~\ref{tab:cross-benchmark} aggregates every
base / SFT cell across all nine benchmark splits and adds DPO and
ORPO runs trained on the identical 100K $(\textsc{chosen},
\textsc{rejected})$ pairs from our pipeline (DPO and ORPO use both
sides of each pair directly; SFT uses only the chosen side). The 7B SFT model improves on \emph{every} cell
($+4.0$ to $+34.4$ EM); 3B SFT records five clear wins
(Delulu, SAFIM-api, SAFIM-blk2, SAFIM-ctl, Real-FIM-Eval), two
near-ties (HE-Inf random-span at parity, SAFIM-blk within $0.1$~EM),
and two regressions on the HumanEval-Infilling Python splits
(single-line $-0.6$, multi-line $-16.1$) --- the splits whose
gold completions most resemble the training-data surface form. The single
most striking cell is SAFIM \textsc{control} 7B, where the base is
nearly broken ($4.8$ EM, ES $0.099$) and SFT recovers it to $39.2$ EM
(ES $0.722$); we attribute the 3B HE-Inf SL/ML regression to a
capacity-vs.-objective trade-off that the 7B has enough slack to
absorb. The mechanism is specific to the multi-line split: the SFT
training set targets single identifier-level FIM holes, so every
curated completion is short --- typically one logical line of code,
even when that line wraps. At 7B, the model retains its pre-trained
ability to generate long, multi-line completions alongside the
anti-hallucination objective it learns from fine-tuning; the two
coexist without interference. At 3B, catastrophic forgetting takes
over: the model over-generalises the ``short completion'' pattern
from training and defaults to abbreviated outputs on any split where
the gold completion spans multiple lines. HumanEval-Infilling
multi-line ($-16.1$ EM at 3B) is precisely that split; single-line
($-0.6$ EM) and random-span ($\pm 0$ EM) are barely affected
because their gold completions are short and already compatible with
what fine-tuning reinforces.

Three patterns separate the contrastive runs from SFT.
\emph{ORPO ties SFT on Delulu} ($+1.0$ EM at 3B, $+0.7$ at 7B) and
is competitive on general FIM at 7B (within $\pm 2$ EM on all eight
cross splits). \emph{DPO collapses across the board}, losing
35--45 EM to SFT on Delulu and broadly 2--49 EM on the cross splits
(the low end is Real-FIM-Eval where absolute numbers are below 5\%);
its $+\!\text{trunc}$ lifts of $+3$ to $+12$ EM on the main cross splits
(vs.\ $\leq 1$ for SFT and ORPO) localise the cause to over-generation
past the FIM-hole boundary --- DPO's contrastive objective does not
internalise the stop.
Two conclusions follow: (i) the synthetic hard negatives are a useful
preference signal that ORPO can exploit; (ii) the chosen side alone,
curated for hard-negative difficulty, matches the strongest contrastive
alternative on the anti-hallucination target and is the safer choice on
general FIM at 3B. We return to the question of when SFT suffices
in \S\ref{sec:analysis:sft}.

\begin{table*}[t]
  \centering\footnotesize
  \setlength{\tabcolsep}{3pt}
  \resizebox{\textwidth}{!}{%
  \begin{tabular}{l cccc | cccc}
  \toprule
   & \multicolumn{4}{c|}{\textbf{3B}} & \multicolumn{4}{c}{\textbf{7B}}\\
  \textbf{Benchmark}      & base & SFT & DPO & ORPO & base & SFT & DPO & ORPO \\
  \midrule
  \multicolumn{9}{l}{\emph{EM (\%, untruncated\,/\,$+\!\text{trunc}$)}} \\
  Delulu                  & 45.9 & 58.7 & 13.3 & \textbf{59.7} & 42.9 & 61.7 & 24.2 & \textbf{62.4} \\
  \quad +trunc            & 52.0 & 59.2 & 16.9 & \textbf{59.9} & 48.8 & 61.8 & 25.8 & \textbf{62.4} \\
  HE-Inf single-line      & 52.4 & 51.8 & 52.4 & \textbf{54.9} & 40.2 & \textbf{57.3} & \phantom{0}8.5 & 56.7 \\
  \quad +trunc            & 52.4 & 51.8 & 48.8 & \textbf{54.9} & 43.3 & \textbf{57.3} & 12.8 & 56.7 \\
  HE-Inf multi-line       & \textbf{27.1} & 11.0 & 19.5 & \phantom{0}9.3 & 16.1 & 22.9 & \phantom{0}3.4 & \textbf{23.7} \\
  \quad +trunc            & \textbf{27.1} & 11.0 & 21.2 & \phantom{0}9.3 & 19.5 & 22.9 & \phantom{0}5.1 & \textbf{23.7} \\
  HE-Inf random-span      & 41.7 & 41.7 & \textbf{47.2} & 40.2 & 34.6 & 44.9 & \phantom{0}8.7 & \textbf{46.5} \\
  \quad +trunc            & \textbf{50.4} & 42.5 & 39.4 & 39.4 & 40.2 & \textbf{47.2} & 21.3 & \textbf{47.2} \\
  SAFIM API               & 50.3 & \textbf{69.0} & 25.5 & 66.8 & 53.9 & \textbf{72.6} & 38.1 & 71.0 \\
  \quad +trunc            & 55.5 & \textbf{69.4} & 26.5 & 66.8 & 58.7 & \textbf{72.6} & 39.7 & 71.0 \\
  SAFIM block             & \textbf{24.6} & 24.5 & 14.6 & 23.9 & 12.4 & \textbf{26.3} & 19.6 & 25.7 \\
  \quad +trunc            & \textbf{30.9} & 24.5 & 19.3 & 23.9 & \textbf{30.1} & 26.3 & 24.7 & 25.8 \\
  SAFIM block-v2          & 27.3 & \textbf{31.1} & 15.3 & \textbf{31.1} & 13.4 & \textbf{32.9} & 22.2 & 32.5 \\
  \quad +trunc            & \textbf{37.5} & 31.1 & 24.5 & 31.1 & \textbf{37.6} & 32.9 & 30.2 & 32.7 \\
  SAFIM control           & 34.7 & \textbf{36.4} & 14.0 & 34.3 & \phantom{0}4.8 & 39.2 & \phantom{0}8.9 & \textbf{41.0} \\
  \quad +trunc            & 35.3 & \textbf{36.4} & 15.0 & 34.3 & \phantom{0}7.6 & 39.3 & 12.0 & \textbf{41.0} \\
  Real-FIM-Eval (add)     & \phantom{0}0.2 & \textbf{3.6}  & 1.2 & 3.3 & \phantom{0}0.2 & \textbf{4.2} & 1.7 & 3.7 \\
  \quad +trunc            & \phantom{0}0.4 & \textbf{3.7}  & 2.3 & 3.3 & \phantom{0}0.4 & \textbf{4.4} & 2.4 & 3.8 \\
  \midrule
  \multicolumn{9}{l}{\emph{ES (untruncated\,/\,$+\!\text{trunc}$)}} \\
  Delulu       & 0.695/0.798 & 0.833/0.835 & 0.288/0.554 & \textbf{0.841}/\textbf{0.841} & 0.628/0.770 & 0.852/0.850 & 0.362/0.624 & \textbf{0.856}/\textbf{0.855} \\
  HE-Inf SL    & 0.815/0.830 & 0.808/0.811 & 0.740/0.633 & \textbf{0.842}/\textbf{0.842} & 0.660/0.767 & \textbf{0.849}/\textbf{0.847} & 0.183/0.596 & 0.848/\textbf{0.847} \\
  HE-Inf ML    & \textbf{0.730}/\textbf{0.761} & 0.581/0.601 & 0.707/0.743 & 0.550/0.564 & 0.487/\textbf{0.715} & 0.690/0.687 & 0.172/0.647 & \textbf{0.694}/0.691 \\
  HE-Inf RS    & 0.702/\textbf{0.810} & 0.723/0.729 & \textbf{0.742}/0.684 & 0.710/0.712 & 0.601/0.757 & 0.745/0.765 & 0.222/0.641 & \textbf{0.770}/\textbf{0.774} \\
  SAFIM API    & 0.667/0.832 & \textbf{0.885}/\textbf{0.884} & 0.370/0.649 & 0.877/0.877 & 0.687/0.838 & \textbf{0.914}/\textbf{0.914} & 0.474/0.715 & 0.908/0.908 \\
  SAFIM blk    & 0.508/\textbf{0.639} & \textbf{0.515}/0.519 & 0.351/0.524 & 0.488/0.490 & 0.285/\textbf{0.621} & \textbf{0.523}/0.525 & 0.405/0.567 & 0.496/0.497 \\
  SAFIM blk2   & 0.514/\textbf{0.663} & \textbf{0.566}/0.568 & 0.348/0.551 & 0.548/0.548 & 0.282/\textbf{0.655} & \textbf{0.580}/0.581 & 0.413/0.597 & 0.560/0.560 \\
  SAFIM ctl    & 0.577/0.667 & \textbf{0.696}/\textbf{0.698} & 0.291/0.532 & 0.695/0.697 & 0.099/0.510 & 0.722/0.723 & 0.163/0.518 & \textbf{0.727}/\textbf{0.727} \\
  Real-FIM-Eval & 0.143/0.185 & 0.318/0.325 & 0.189/\textbf{0.336} & \textbf{0.329}/0.333 & 0.140/0.182 & 0.331/0.339 & 0.193/0.340 & \textbf{0.338}/\textbf{0.343} \\
  \bottomrule
  \end{tabular}}
  \caption{EM (\%) and ES across all nine FIM benchmark splits for the three
  fine-tuning recipes (SFT, DPO, ORPO) and the base model, at 3B and
  7B. Bold marks the best cell per (size, benchmark) row for each metric.
  SFT and ORPO are within $\pm 2$ EM on most cells; DPO loses
  broadly 7--49 EM to both on the cross splits. The +trunc rows truncate
  predictions to the gold completion's line count: trunc lifts the
  base by up to $+25$~EM (SAFIM blk2, 7B) and lifts DPO by $+3$ to
  $+12$~EM on cross splits, but moves SFT and ORPO cells by $\leq\!1$~EM,
  showing that SFT and ORPO internalise the FIM-hole stop while DPO
  and the base over-generate. On Real-FIM-Eval ES, ORPO edges out SFT
  at 7B (0.338/0.343 vs.\ 0.331/0.339) while DPO's +trunc ES
  (0.336) leads at 3B, suggesting contrastive training improves
  edit-level proximity on free-form edits even when EM lags.}  \label{tab:cross-benchmark}
\end{table*}

\subsection{Ablations}
\label{sec:experiments:ablations}

The proof of concept uses one fixed recipe: 100K curated rows, all
four hallucination types, all eight source languages, 7B base. We
isolate each design choice and confirm that the conclusions are
not 7B-specific. Every ablation cell shares the exact same training
hyperparameters and evaluation protocol
(\S\ref{sec:experiments:setup}); only the curator's row selector
varies. We evaluate every cell on a fixed six-benchmark panel that
spans the main shapes of FIM tasks (single-line Python, multi-line
Python, random-span Python, API-completion, control-flow, and the
multilingual Delulu benchmark itself). Both EM and ES are reported
for each ablation cell; cells are formatted as
untruncated\,/\,$+\!\text{trunc}$ pairs.
to preserve the stopping signal in one number; truncation moves
trained cells by $\leq 4$~EM throughout, consistent with
\S\ref{sec:experiments:poc}. Table~\ref{tab:ablations} aggregates all
four axes; we discuss each in turn.

\begin{table*}[t]
  \centering\footnotesize
  \setlength{\tabcolsep}{3pt}
  \begin{tabular}{l cccccc}
  \toprule
  \textbf{Cell} & \textbf{Delulu} & \textbf{HE-SL} & \textbf{HE-ML} & \textbf{HE-RS} & \textbf{SA-api} & \textbf{SA-ctl} \\
  \midrule
  \multicolumn{7}{l}{\emph{EM (\%, untruncated\,/\,$+\!\text{trunc}$)}} \\
  \multicolumn{7}{l}{\emph{(a) Size axis, 7B; rows $\in\{5\text{K},20\text{K},50\text{K},100\text{K}\}$}} \\
  \phantom{0}5{,}000      & 18.7/18.7 & 25.6/26.8 & \phantom{0}3.4/\phantom{0}4.2 & 21.3/23.6 & 40.3/40.3 & \phantom{0}4.8/\phantom{0}5.1 \\
  20{,}000                & 26.9/27.0 & 43.3/43.9 & 19.5/20.3 & 38.6/40.9 & 52.9/55.5 & 11.8/11.9 \\
  50{,}000                & \textbf{61.7}/\textbf{61.8} & \textbf{56.7}/\textbf{56.7} & \textbf{27.1}/\textbf{27.1} & \textbf{48.0}/\textbf{50.4} & \textbf{72.6}/\textbf{72.6} & \textbf{36.8}/\textbf{36.9} \\
  \multicolumn{7}{l}{\emph{(b) Type axis, 7B, 32K rows per single-type cell}} \\
  \textsc{import}         & 25.5/25.5 & \textbf{55.5}/\textbf{55.5} & \phantom{0}0.8/\phantom{0}0.8 & 34.6/34.6 & 15.8/15.8 & \phantom{0}0.4/\phantom{0}0.4 \\
  \textsc{method}         & \textbf{57.3}/\textbf{57.5} & 52.4/53.7 & 24.6/24.6 & \textbf{40.2}/\textbf{48.0} & \textbf{71.6}/\textbf{72.3} & 24.1/25.5 \\
  \textsc{parameter}      & 57.0/57.3 & 50.6/54.3 & \textbf{25.4}/\textbf{26.3} & 39.4/46.5 & 70.3/70.3 & 24.0/24.9 \\
  \textsc{undef.\ var.}   & 57.1/57.2 & 51.2/53.0 & 22.9/25.4 & \textbf{40.2}/47.2 & 70.6/71.0 & \textbf{24.9}/\textbf{25.8} \\
  \multicolumn{7}{l}{\emph{(c) Language axis, 7B, 16K rows per cell}} \\
  Python only             & \phantom{0}7.8/\phantom{0}8.1 & 20.1/20.1 & \phantom{0}6.8/\phantom{0}6.8 & 14.2/15.7 & 28.4/28.4 & \phantom{0}3.0/\phantom{0}3.2 \\
  top-5 langs             & \textbf{31.9}/\textbf{33.0} & \textbf{55.5}/\textbf{55.5} & \textbf{27.1}/\textbf{27.1} & 40.9/\textbf{46.5} & \textbf{53.5}/\textbf{57.1} & \textbf{15.8}/\textbf{15.9} \\
  all 8 langs             & 31.1/31.8 & 53.0/53.0 & 22.0/22.0 & \textbf{41.7}/44.9 & 50.6/53.9 & 10.2/10.3 \\
  \multicolumn{7}{l}{\emph{(d) Re-runs at 3B (one cell per axis)}} \\
  size-20K                & 53.2/53.6 & 50.0/50.0 & 13.6/14.4 & 40.2/43.3 & 65.8/66.1 & 15.0/15.0 \\
  type-method             & 52.5/53.0 & 50.0/50.6 & \textbf{21.2}/\textbf{21.2} & \textbf{41.7}/\textbf{44.1} & 67.4/67.7 & 19.7/19.7 \\
  lang-top5               & 48.6/49.0 & 42.7/42.7 & 11.9/12.7 & 40.2/42.5 & 64.5/64.8 & 12.8/12.9 \\
  recipe (3B, 100K)       & \textbf{58.7}/\textbf{59.2} & \textbf{51.8}/\textbf{51.8} & 11.0/11.0 & \textbf{41.7}/42.5 & \textbf{69.0}/\textbf{69.4} & \textbf{36.4}/\textbf{36.4} \\
  \midrule
  \multicolumn{7}{l}{\emph{Reference: 7B 100K recipe (all types, all 8 langs)}} \\
  recipe (7B, 100K)       & 61.7/61.8 & 57.3/57.3 & 22.9/22.9 & 44.9/47.2 & 72.6/72.6 & 39.2/39.3 \\
  \midrule
  \multicolumn{7}{l}{\emph{ES (untruncated\,/\,$+\!\text{trunc}$)}} \\
  \multicolumn{7}{l}{\emph{(a) Size axis, 7B}} \\
  \phantom{0}5{,}000      & 0.605/0.689 & 0.481/0.649 & 0.200/0.613 & 0.385/0.626 & 0.759/0.816 & 0.198/0.449 \\
  20{,}000                & 0.591/0.622 & 0.731/0.756 & 0.618/0.695 & 0.702/0.752 & 0.706/0.783 & 0.483/0.537 \\
  50{,}000                & \textbf{0.849}/\textbf{0.850} & \textbf{0.842}/\textbf{0.841} & \textbf{0.731}/\textbf{0.727} & \textbf{0.761}/\textbf{0.782} & \textbf{0.915}/\textbf{0.915} & \textbf{0.711}/\textbf{0.713} \\
  \multicolumn{7}{l}{\emph{(b) Type axis, 7B}} \\
  \textsc{import}         & 0.570/0.570 & \textbf{0.823}/0.823 & 0.371/0.371 & 0.623/0.623 & 0.599/0.600 & 0.339/0.339 \\
  \textsc{method}         & \textbf{0.812}/\textbf{0.815} & 0.802/\textbf{0.842} & 0.604/0.738 & 0.699/0.783 & 0.892/\textbf{0.907} & 0.585/\textbf{0.655} \\
  \textsc{parameter}      & 0.809/0.810 & 0.781/0.837 & \textbf{0.640}/\textbf{0.745} & 0.696/\textbf{0.788} & \textbf{0.894}/0.901 & \textbf{0.594}/0.653 \\
  \textsc{undef.\ var.}   & 0.808/0.810 & 0.795/0.839 & 0.583/0.739 & \textbf{0.707}/0.786 & 0.894/0.903 & 0.591/0.651 \\
  \multicolumn{7}{l}{\emph{(c) Language axis, 7B}} \\
  Python only             & 0.317/0.385 & 0.508/0.581 & 0.359/0.550 & 0.442/0.563 & 0.431/0.502 & 0.307/0.420 \\
  top-5 langs             & \textbf{0.639}/\textbf{0.664} & \textbf{0.830}/\textbf{0.833} & \textbf{0.710}/\textbf{0.739} & \textbf{0.725}/\textbf{0.778} & \textbf{0.772}/\textbf{0.822} & \textbf{0.512}/\textbf{0.562} \\
  all 8 langs             & 0.604/0.628 & 0.809/0.818 & 0.640/0.720 & 0.717/0.768 & 0.712/0.758 & 0.432/0.493 \\
  \multicolumn{7}{l}{\emph{(d) Re-runs at 3B}} \\
  size-20K                & 0.790/0.795 & 0.754/0.786 & 0.573/0.644 & 0.711/\textbf{0.753} & 0.857/0.868 & 0.455/0.533 \\
  type-method             & 0.764/0.770 & 0.695/0.727 & \textbf{0.623}/\textbf{0.678} & 0.669/0.697 & 0.875/0.882 & 0.464/0.524 \\
  lang-top5               & 0.755/0.761 & 0.699/0.754 & 0.518/0.633 & 0.673/0.748 & 0.854/0.868 & 0.417/0.519 \\
  recipe (3B, 100K)       & \textbf{0.833}/\textbf{0.835} & \textbf{0.808}/\textbf{0.811} & 0.581/0.601 & \textbf{0.723}/0.729 & \textbf{0.885}/\textbf{0.884} & \textbf{0.696}/\textbf{0.698} \\
  \midrule
  \multicolumn{7}{l}{\emph{Reference: 7B 100K recipe (all types, all 8 langs)}} \\
  recipe (7B, 100K)       & 0.852/0.850 & 0.849/0.847 & 0.690/0.687 & 0.745/0.765 & 0.914/0.914 & 0.722/0.723 \\
  \bottomrule
  \end{tabular}
  \caption{Ablation cells (EM and ES, formatted as
  untruncated\,/\,$+\!\text{trunc}$). Groups (a)--(c) vary one
  curator axis at a time at 7B; group (d) re-runs one representative
  cell per axis at 3B.
  Bold marks the within-axis best per metric (excluding the reference
  recipe row). Across all four groups, $+\!\text{trunc}$ moves trained
  cells by $\leq 4$~EM: the stopping behaviour learned in
  \S\ref{sec:experiments:poc} is invariant to the recipe axis varied.}
  \label{tab:ablations}
\end{table*}

\subsubsection{Size: how many rows are enough?}
\label{sec:abl:size}

Uniformly subsample the 100K curated pool to $\{5\text{K}, 20\text{K},
50\text{K}, 100\text{K}\}$; per-bucket and per-repo caps unchanged,
language and type mix unchanged. All five benchmarks share the same
saturating-step shape (Table~\ref{tab:ablations}a, plus the 100K
reference row): 5K is far below where the recipe works at all on
Delulu and SAFIM-control; 20K is mid-curve; \emph{the knee falls
between 20K and 50K rows on every benchmark}; 50K and 100K are within
noise, with HE-RS even mildly regressing. The recipe is therefore not
relying on the full 100K; 50K would have produced a near-identical
paper headline. The shared curve shape across very different
benchmarks suggests the 50K-row ceiling is a property of the
\emph{recipe}, not of any one benchmark.

\subsubsection{Hallucination type: are the four types interchangeable?}
\label{sec:abl:type}

Four single-type 7B cells, each trained on $\sim$32K rows from one
taxonomy type over the same eight languages; budget, caps,
hyperparameters and language mix held fixed. \textsc{Method},
\textsc{parameter}, and \textsc{undefined-variable} produce
near-identical numbers on every benchmark (within $\sim 2$ EM of each
other, Table~\ref{tab:ablations}b): training on any one of these
three transfers to the others through a shared ``do not invent
identifiers'' signal. \textsc{Import-only} breaks the pattern: best
on its own Delulu type but \emph{catastrophically} loses on
multi-line completions (HE-ML $0.8$, SAFIM-ctl $0.4$) and API
completion (SAFIM-api $15.8$). Imports are short and lexically
distinctive, so an import-only model learns a degenerate ``emit a
short import-shaped prefix'' policy that crashes when the gold
completion is a longer block. The all-types 100K recipe matches or
beats every single-type cell on every benchmark.

\subsubsection{Language coverage: breadth vs.\ budget}
\label{sec:abl:lang}

\begin{figure}[t]
  \centering
  \includegraphics[width=0.65\textwidth]{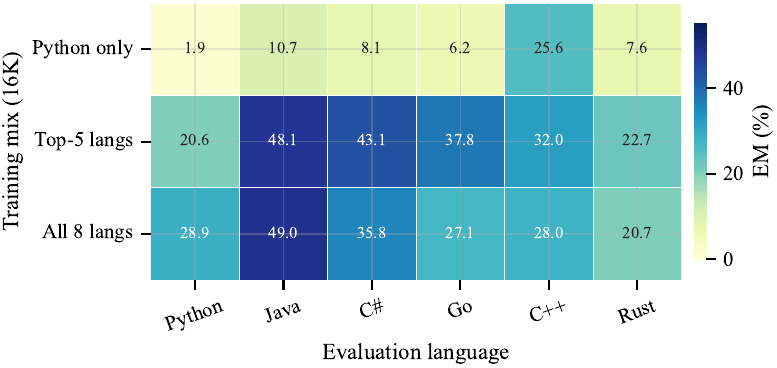}
  \caption{Language ablation, 7B. Per-evaluation-language Delulu EM
  at a fixed 16K-row training budget. Python rises from 1.9 to 28.9
  EM as we add non-Python languages, none of which include any Python
  data; once $\ge\!5$ diverse languages are present at this budget,
  adding more breadth no longer helps overall.}
  \label{fig:abl-lang}
\end{figure}

Three 16K-row 7B cells: \textsc{lang-python} (Python only),
\textsc{lang-top5} (the five largest non-Python languages: C\#, Go,
Java, JavaScript, PHP), and \textsc{lang-all} (all eight training
languages, uniform per-bucket sampling). At 16K, Python-only
collapses to $7.8$ EM on the multilingual Delulu val
\emph{and even to $1.9$ EM on Python itself}: squeezing a small
budget into one language distorts the multilingual prior so badly
that the trained language gets hurt. Replacing the 16K Python rows
with the same 16K rows drawn from five non-Python languages
quadruples Delulu ($7.8 \to 32.0$ EM) and \emph{raises Python to
$20.6$ EM without a single Python row in training}
(Figure~\ref{fig:abl-lang}). Adding the remaining three languages
(\textsc{lang-all}) leaves overall Delulu within noise of top-5
($31.1$ vs.\ $32.0$) and raises Python further to $28.9$. At the
same 16K budget, top-5 and all-8 are tied
(Table~\ref{tab:ablations}c); the remaining gap to the 100K recipe
is \emph{scale}, not breadth. The recipe needs roughly five diverse
languages to ignite cross-lingual transfer; beyond that, budget is
better spent scaling up than diversifying further.

The cross-lingual effect is best understood by noting that all four
hallucination types target \emph{identifier-level} edits: replacing a
method name, fabricating an argument, introducing an undefined variable,
or swapping an import line. These edit patterns are structurally
language-independent --- a plausible-but-wrong method call looks the
same in Go as it does in Python --- so the model's identifier-level
discrimination signal generalises across language boundaries even when
no Python rows appear in training.  The five-language threshold likely
marks the point at which the training set covers enough surface-form
diversity (camelCase, snake\_case, module paths, different import
syntaxes) for this shared discriminative structure to transfer.

\subsubsection{Does the ablation behaviour transfer to 3B?}
\label{sec:abl:size3b}

We re-run one cell per axis at 3B (Table~\ref{tab:ablations}d). The
same qualitative pattern holds: the 100K recipe is best on four of
five benchmarks; the size-20K cell is consistently below the 100K
cell; the lang-top5 cell loses ground on Delulu but stays close on
HE-Inf single-line. HE-Inf multi-line remains the single anomaly:
best at type-method, not at the full recipe, in line with the
size-dependent trade-off observed in the proof of concept.

\subsubsection{Does the recipe transfer across base model families?}
\label{sec:abl:basemodels}

\emph{Setup.} The proof of concept fine-tunes
\textsc{Qwen2.5-Coder-7B-Instruct}. To check that the curated dataset
is the load-bearing piece --- and not an artifact of Qwen's tokenizer,
chat template, or pre-training mixture --- we re-train two additional
7B base models on the \emph{exact same} 100K curated rows:
\textsc{StarCoder2-7B} and \textsc{CodeLlama-7b-hf}. Both are
non-instruct FIM-pretrained bases with different vocabularies, FIM
token conventions (PSM vs.\ SPM), and pre-training corpora.

\emph{Same.} Training hyperparameters
(\S\ref{sec:experiments:setup}), the 100K-row dataset, the eight
evaluation benchmarks, and the EM/ES scoring pipeline
are identical across the three families. The only per-family
adjustment is the FIM prompt template: each base model is fed its
native \texttt{<fim\_*>} sentinels, so the curated rows are
retokenised --- but never re-curated --- for StarCoder2 and CodeLlama.

\emph{Changed.} The base model family
(Qwen-instruct $\to$ StarCoder2-base $\to$ CodeLlama-base) and, by
necessity, the FIM template.

\begin{figure}[t]
  \centering
  \includegraphics[width=0.65\textwidth]{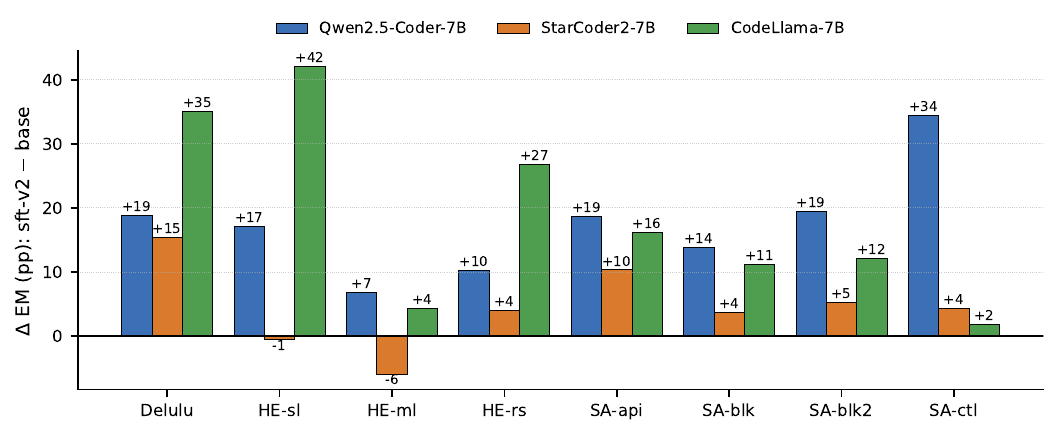}
  \caption{Per-benchmark EM lift ($\Delta\,\text{EM} =
  \text{sft-v2}-\text{base}$, percentage points) for the three 7B
  base model families trained on the identical 100K curated dataset.
  Lifts are positive on $23/24$ cells; the only regression is
  StarCoder2 on HE-ml ($-6.0$ pp), where the base is already strong
  and ES barely moves (Table~\ref{tab:abl-basemodels}). The weaker
  the base on a cell, the larger the lift --- the recipe rescues
  CodeLlama's near-empty HumanEval-Infilling single-/random-line and
  Qwen's collapsed SAFIM control-flow, while leaving StarCoder2's
  already-strong HE cells essentially intact.}
  \label{fig:abl-basemodels}
\end{figure}

\emph{Result.} The recipe lifts \emph{every} (family, benchmark)
cell on both EM and ES except StarCoder2's HE-sl/ml,
where SFT preserves rather than improves an already-strong base
(Table~\ref{tab:abl-basemodels}, Figure~\ref{fig:abl-basemodels}).
Three observations stand out. First, the Delulu lift is uniformly
large across families ($+15$ to $+35$ EM points; ES from
$\sim\!0.41$--$0.66$ to $\sim\!0.80$--$0.85$), confirming that the
hallucination-correction signal is encoded in the data rather than in
the base model. Second, the largest relative gain is on
\textsc{CodeLlama}'s HumanEval-Infilling single-line: $7.3 \to 49.4$
EM ($0.228 \to 0.718$ ES); the base barely emits a well-formed FIM
completion on this split, and the curated rows teach it to do so.
Third, Qwen sees a striking SAFIM
control-flow lift ($+34.4$ EM, ES $0.099 \to 0.722$): SAFIM-ctl is
the cell where the Qwen base was weakest on this panel, and the
curated rows close most of that gap. StarCoder2, whose base is
already the strongest of the three on HumanEval-Infilling, shows the
smallest gains there (HE-ml slips by $-6.0$ EM but its ES drops only
$-0.037$, indicating shorter but still close completions), while
still gaining $+15$ EM on Delulu and $+4$--$10$ EM on the SAFIM
panel. Reading the +trunc rows of Table~\ref{tab:abl-basemodels}
makes the data effect even sharper: the EM gap between base and SFT
shrinks once the base is allowed to be truncated to gold line count,
but the SFT models still win on every cell and their ES is essentially
unchanged before/after truncation. The recipe therefore behaves as a
base-model-agnostic post-training step that teaches both the right
completion \emph{and} the right stopping point.

\begin{table*}[t]
  \centering\footnotesize
  \setlength{\tabcolsep}{3pt}
  \begin{tabular}{llcccccccc}
  \toprule
  \textbf{Model} & \textbf{Run}
    & \textbf{Delulu} & \textbf{HE-sl} & \textbf{HE-ml} & \textbf{HE-rs}
    & \textbf{SA-api} & \textbf{SA-blk} & \textbf{SA-blk2} & \textbf{SA-ctl} \\
  \midrule
  \multicolumn{10}{l}{\emph{EM (\%)}} \\
  Qwen2.5-Coder-7B-Instr. & base          & 42.9 & 40.2 & 16.1 & 34.6 & 53.9 & 12.4 & 13.4 & \phantom{0}4.8 \\
                          & \quad +trunc  & 48.8 & 43.3 & 19.5 & 40.2 & 58.7 & 30.1 & 37.6 & \phantom{0}7.6 \\
                          & sft-v2        & \textbf{61.7} & \textbf{57.3} & \textbf{22.9} & \textbf{44.9} & \textbf{72.6} & \textbf{26.3} & 32.9 & \textbf{39.2} \\
                          & \quad +trunc  & \textbf{61.8} & \textbf{57.3} & \textbf{22.9} & \textbf{47.2} & \textbf{72.6} & \textbf{26.3} & 32.9 & \textbf{39.3} \\
  StarCoder2-7B           & base          & 43.9 & 51.8 & \textbf{24.6} & 38.6 & 57.7 & 26.1 & 31.1 & 36.3 \\
                          & \quad +trunc  & 48.2 & 53.0 & \textbf{24.6} & 45.7 & 61.9 & 32.7 & \textbf{39.3} & 36.8 \\
                          & sft-v2        & \textbf{59.3} & 51.2 & 18.6 & 42.5 & \textbf{68.1} & \textbf{29.8} & 36.3 & \textbf{40.6} \\
                          & \quad +trunc  & \textbf{59.6} & 51.2 & 18.6 & 42.5 & \textbf{68.7} & \textbf{29.8} & 36.3 & \textbf{40.6} \\
  CodeLlama-7b-hf         & base          & 23.6 & \phantom{0}7.3 & \phantom{0}6.8 & \phantom{0}7.9 & 45.5 & \phantom{0}7.2 & 10.6 & 24.9 \\
                          & \quad +trunc  & 46.2 & 10.4 & \phantom{0}9.3 & 19.7 & 58.4 & 28.8 & 36.0 & 27.2 \\
                          & sft-v2        & \textbf{58.7} & \textbf{49.4} & \textbf{11.0} & \textbf{34.6} & \textbf{61.6} & \textbf{18.3} & \textbf{22.7} & \textbf{26.7} \\
                          & \quad +trunc  & \textbf{58.8} & \textbf{49.4} & \textbf{11.0} & \textbf{34.6} & \textbf{61.9} & \textbf{19.4} & \textbf{23.6} & \textbf{26.8} \\
  \midrule
  \multicolumn{10}{l}{\emph{ES}} \\
  Qwen2.5-Coder-7B-Instr. & base          & 0.628 & 0.660 & 0.487 & 0.601 & 0.687 & 0.285 & 0.282 & 0.099 \\
                          & \quad +trunc  & 0.770 & 0.767 & 0.715 & 0.757 & 0.838 & 0.621 & 0.655 & 0.510 \\
                          & sft-v2        & \textbf{0.852} & \textbf{0.849} & \textbf{0.690} & \textbf{0.745} & \textbf{0.914} & 0.523 & 0.580 & \textbf{0.722} \\
                          & \quad +trunc  & \textbf{0.850} & \textbf{0.847} & \textbf{0.687} & \textbf{0.765} & \textbf{0.914} & 0.525 & 0.581 & \textbf{0.723} \\
  StarCoder2-7B           & base          & 0.658 & \textbf{0.773} & \textbf{0.711} & 0.670 & 0.753 & 0.555 & 0.580 & 0.609 \\
                          & \quad +trunc  & 0.731 & \textbf{0.799} & \textbf{0.731} & \textbf{0.773} & \textbf{0.877} & \textbf{0.664} & \textbf{0.686} & 0.687 \\
                          & sft-v2        & \textbf{0.830} & 0.754 & 0.674 & 0.716 & \textbf{0.881} & \textbf{0.610} & \textbf{0.647} & \textbf{0.711} \\
                          & \quad +trunc  & \textbf{0.833} & 0.758 & 0.674 & 0.722 & \textbf{0.885} & 0.624 & 0.656 & \textbf{0.716} \\
  CodeLlama-7b-hf         & base          & 0.409 & 0.228 & 0.416 & 0.237 & 0.602 & 0.254 & 0.259 & 0.446 \\
                          & \quad +trunc  & 0.727 & 0.465 & 0.625 & 0.584 & 0.840 & 0.627 & 0.651 & 0.619 \\
                          & sft-v2        & \textbf{0.804} & \textbf{0.718} & \textbf{0.562} & \textbf{0.623} & \textbf{0.849} & \textbf{0.450} & \textbf{0.473} & \textbf{0.565} \\
                          & \quad +trunc  & \textbf{0.809} & \textbf{0.725} & \textbf{0.569} & \textbf{0.630} & \textbf{0.850} & \textbf{0.466} & \textbf{0.481} & \textbf{0.567} \\
  \bottomrule
  \end{tabular}
  \caption{Same 100K curated dataset applied to three 7B base model
  families. Each (family, benchmark) cell improves on both EM and
  ES except StarCoder2's HE-sl/ml, where SFT preserves
  rather than improves an already-strong base. The +trunc rows show
  why the EM lifts in this table understate the data effect: the base
  models over-generate, so truncating to gold-line-count rescues a
  large chunk of base EM (e.g.\ Qwen SAFIM-blk $12.4\to 30.1$, CodeLlama
  Delulu $23.6\to 46.2$), while the SFT cells barely move ($\leq\!1$~EM
  on every cell). The SFT recipe therefore teaches \emph{both} the
  right token sequence \emph{and} when to stop; over the base, ES
  remains the cleaner one-number summary. $N$ per benchmark: Delulu
  $1950$, HE-sl $164$, HE-ml $118$, HE-rs $127$, SA-api $310$, SA-blk
  $8781$, SA-blk2 $4571$, SA-ctl $8629$.}
  \label{tab:abl-basemodels}
\end{table*}

\subsubsection{Difficulty-aware curation: fool-rate threshold}
\label{sec:experiments:ablations:foolrate}

The previous ablations vary what the curator keeps (size, type,
language).  This axis varies \emph{how hard} the kept rows are.
Each curated row carries a \emph{fool rate}~$\tau$ --- the fraction
of a three-judge panel (\textsc{GPT-4o-mini}, \textsc{GPT-4.1-mini}, \textsc{GPT-5.4-mini}) that preferred the hard negative over the
golden completion (\S\ref{sec:method:fool-rate}).  We partition the
curated pool into three non-overlapping buckets ---
\textsc{easy} ($\tau \le 0.33$),
\textsc{medium} ($0.33 < \tau \le 0.66$), and
\textsc{hard} ($\tau > 0.66$) --- and draw a balanced subset of
5{,}000 rows from each, matching per-(type, language) counts
exactly so that the only variable is fool-rate difficulty.  We
deliberately shrink the bucket size from earlier 16{,}589-row runs:
at the larger scale all three buckets converged to within
$<\!1$~EM point, saturating any difficulty signal.  At 5K the
recipe is still informative but no longer saturated, exposing
how much each judge-defined tier contributes per row.  We focus
on Qwen2.5-Coder-7B here.

\paragraph{Result.}
Table~\ref{tab:fool-rate-ablation} shows a clear, monotone
difficulty effect: overall Delulu~EM rises from
$32.7$ (\textsc{easy}) to $37.4$ (\textsc{medium}) to
$37.7$ (\textsc{hard}) --- a $\sim\!5$ point lift from easy to
medium and a near-plateau from medium to hard.  The plateau is
informative: once the bucket reaches the panel-disagreement
regime ($\tau > 0.33$), additional ``trickiness'' adds little.
The supervisory mass lives in the disputed-but-not-impossible
middle and tail, not in the unanimous-easy tier.

Three patterns are worth noting.  First, \textsc{import} drives
most of the lift: it climbs monotonically on both metrics from
$36.9 \to 43.2 \to 44.9$~EM ($+8.0$ pp) and
$0.707 \to 0.745 \to 0.757$~ES ($+0.050$),
consistent with the earlier 16{,}589-row run where
\textsc{import} was also the one type that responded to harder
negatives.  Import-name hallucinations require sharper
discriminative signal than the other three types.
Second, \textsc{method}, \textsc{parameter}, and \textsc{undef.}
all flatten between medium and hard on EM (within $\pm\!0.7$),
suggesting that for these types the medium bucket already
contains enough disagreement to teach the contrast and the
hard tail offers diminishing returns.
Third, ES tells a subtler story than EM on these three types:
ES peaks at medium and \emph{dips} at hard
(method $0.714 \to 0.689$, param.\ $0.710 \to 0.689$,
undef.\ $0.714 \to 0.697$), while EM stays flat.  Hardest
negatives sharpen the binary golden/hallucinated decision but
pull surface forms away from the gold token sequence --- the
model still picks the right behavior, just expressed less like
the reference string.  Net: the supervisory mass for non-import
types lives in the disputed middle, not the unanimous-hard tail.

\emph{Caveat.}
Delulu contains fewer than 50 samples for several (type, language)
cells; the per-type and per-language breakdowns should therefore be
read as suggestive rather than definitive.

\begin{table}[t]
  \centering\footnotesize
  \setlength{\tabcolsep}{4pt}
  \begin{tabular}{l cc cc cc}
  \toprule
  & \multicolumn{2}{c}{\textsc{easy}} & \multicolumn{2}{c}{\textsc{medium}} & \multicolumn{2}{c}{\textsc{hard}} \\
  \cmidrule(lr){2-3}\cmidrule(lr){4-5}\cmidrule(lr){6-7}
  & EM & ES & EM & ES & EM & ES \\
  \midrule
  \emph{Overall}
                  & 32.7 & 0.692 & 37.4 & \textbf{0.721} & \textbf{37.7} & 0.708 \\
  \midrule
  \multicolumn{7}{l}{\emph{By hallucination type}} \\
  \textsc{import} & 36.9 & 0.707 & 43.2 & 0.745 & \textbf{44.9} & \textbf{0.757} \\
  \textsc{method} & 30.4 & 0.692 & \textbf{34.3} & \textbf{0.714} & \textbf{34.3} & 0.689 \\
  \textsc{param.} & 33.3 & 0.683 & \textbf{37.5} & \textbf{0.710} & 36.8 & 0.689 \\
  \textsc{undef.} & 30.5 & 0.685 & 35.0 & \textbf{0.714} & \textbf{35.2} & 0.697 \\
  \bottomrule
  \end{tabular}
  \caption{Fool-rate threshold ablation (Qwen2.5-Coder-7B).  Each
  bucket contains 5{,}000 rows drawn from the curated pool with
  identical per-(type, language) distributions; the only variable
  is the fool-rate difficulty.  At this smaller scale the
  difficulty signal becomes visible: harder negatives lift
  Delulu~EM by $\sim\!5$~points
  (easy~32.7 $\to$ medium~37.4 $\to$ hard~37.7), driven mainly by
  \textsc{import} (+8.0~EM, +0.050~ES).  On the other three types
  EM plateaus between medium and hard while ES actually peaks at
  medium and dips at hard --- harder negatives sharpen the
  golden/hallucinated decision boundary but pull surface forms
  away from the gold token sequence, indicating that the
  supervisory mass sits in the disputed middle, not the
  unanimous-easy or unanimous-hard tier.  Bold marks the
  within-row best per metric.  $N\!=\!1{,}950$ (Delulu val);
  per-type $N$ ranges from 435 to 577.}
  \label{tab:fool-rate-ablation}
\end{table}

%% file: sections/analysis.tex
\section{Analysis}
\label{sec:analysis}

The proof-of-concept results
(\S\ref{sec:experiments:poc}) and ablations
(\S\ref{sec:experiments:ablations}) already integrate the per-type,
per-language, size-scaling and language-coverage analyses next to
the corresponding tables. This section focuses on three cross-cutting
topics: the size-dependent trade-off at 3B
(\S\ref{sec:analysis:size}), when SFT on hard negatives is sufficient
without explicit contrastive training (\S\ref{sec:analysis:sft}),
and failure modes encountered during pipeline development
(\S\ref{sec:analysis:failures}).

\subsection{Size-dependent trade-off}
\label{sec:analysis:size}

The clearest cross-benchmark pattern in Table~\ref{tab:cross-benchmark}
is the contrast between 3B and 7B. On Delulu both sizes improve
substantially (+12.8 and +18.8 EM); on multilingual SAFIM the 7B
recovers a near-broken \textsc{control} subset (+34.4) while the 3B
gain is muted (+1.7). The most informative diagnostic, however, is
the 3B regression on HumanEval-Infilling single- and multi-line: the
two splits where the base 3B is at its strongest, and where the gold
completions are short, idiomatic Python.

We interpret this as a capacity-versus-objective trade-off. The SFT
objective asks the model to prefer the (often longer, multilingual)
golden completion over plausible identifier-level hard negatives; at 3B
there is little slack between the base FIM distribution and the
direction the fine-tuning task pulls it, and the optimiser pays for
hallucination defence with a fraction of general-FIM accuracy on the
splits closest to the base's strengths. At 7B there is enough
capacity to satisfy both. Two pieces of evidence support this reading.
\emph{(i)} HumanEval-Infilling random-span --- the split that most
resembles arbitrary FIM holes rather than the canonical short Python
shape --- is the only HumanEval-Infilling subset where the 3B does
not regress. \emph{(ii)} The size ablation (Table~\ref{tab:ablations}a)
shows that even at 3B the recipe behaves monotonically with budget;
the regression is not the recipe getting worse with more data, it is
a fixed bias in the SFT-target distribution that 3B has too little
spare capacity to absorb. The fool-rate-threshold
ablation (\S\ref{sec:experiments:ablations:foolrate}), run at 7B
on smaller 5K buckets that no longer saturate the difficulty
signal, shows the complementary picture for the data side: harder
buckets lift Delulu EM by $\sim\!5$~points (easy~32.7 $\to$
medium~37.4 $\to$ hard~37.7), so the supervisory mass lives in the
disputed-middle and tail of the fool-rate distribution rather than
in the unanimous-easy tier.

\subsection{When does SFT suffice?}
\label{sec:analysis:sft}

The SFT vs.\ DPO vs.\ ORPO comparison (Table~\ref{tab:cross-benchmark},
\S\ref{sec:experiments:poc}) surfaces a practically important finding
that deserves emphasis: \emph{training on only the chosen side of a
preference pair matches, and sometimes exceeds, the best contrastive
alternative when the training distribution has been curated for
hard-negative difficulty}.  ORPO, which sees both the chosen and
rejected completions, ties SFT on the anti-hallucination target
($\pm 1$ EM, $\pm 0.004$ ES across all benchmarks) and is competitive
on general FIM at 7B.  DPO, using the same pairs, collapses.

Two properties of our setting explain the gap.  First, the hard
negatives are already in the right difficulty regime: generated by
frontier models and scored by a judge panel, they are neither trivially
distinguishable from the gold nor adversarially far from it.  A
well-curated SFT target set therefore carries most of the preference
information that an explicit contrastive objective would add.  Second,
DPO's margin-based objective is sensitive to over-long generations,
and FIM is a stopping-sensitive task; SFT's cross-entropy loss on the
short, boundary-respecting golden completions aligns naturally with
what the task requires, while DPO's gradient can push toward any
completion that differs from the rejected sequence --- including ones
that over-generate past the FIM hole.  These results suggest that
for \emph{stopping-sensitive FIM tasks}, a high-quality curated SFT
corpus with hard-negative selection is a safer default than paired
contrastive training.

\subsection{Failure modes during pipeline development}
\label{sec:analysis:failures}

\paragraph{(1) Plausibility is a knob.} A naive prompt template lets
the generator emit tokens with explicit ``hallucinated'' tells such as
\texttt{Fake\dots}, \texttt{Bogus\dots}, or \texttt{Hallucinated\dots}.
Such rows are trivial supervised signal: the model learns to suppress
the surface artifact rather than the underlying mistake. We instruct
the generator to keep new identifiers \emph{plausible}, which both
makes the SFT signal more useful and, as the fool-rate ablation
confirms (\S\ref{sec:experiments:ablations:foolrate}), increases how
often the generated row deceives the judge panel.

\paragraph{(2) Position matters more than content for the import
type.} An early version of the pipeline placed the candidate import at
the end of the prefix --- the same hole position used for the other
three types --- which made the candidate appear \emph{after} its first
usage and silently corrupted the supervisory signal. We restructure
the FIM hole so it coincides with the line of the original import
(Appendix~\ref{app:implementation:import-restructure}). The general
principle: any extension of the taxonomy must place the candidate at
the location it would naturally occupy in the file.

%% file: sections/limitations.tex
\section{Limitations}
\label{sec:limitations}

\paragraph{Synthetic vs.\ natural hallucinations.} Hallucinated
completions in our training set are produced by adversarial prompting
of strong generators rather than collected from in-the-wild Code-LLM
mistakes. Some real-world hallucinations (off-by-one errors,
async/await misuse, type-annotation drift) are not covered by our
four-type taxonomy. We expect the recipe to extend cleanly to new
types by writing a new generator prompt
(Appendix~\ref{app:prompts:generator}), but that claim is currently
untested.

\paragraph{Real-FIM-Eval absolute performance.}
The SFT model achieves only $3.6$--$4.2$ EM on Real-FIM-Eval (Table~\ref{tab:cross-benchmark}),
versus $60$+ EM on Delulu.  Three factors explain this.  First, the
Real-FIM-Eval (add) split samples arbitrary real developer edits from
228 permissively-licensed repositories; many of these edits are
multi-token insertions that fall outside the identifier-level
hallucination regime our recipe targets.  Second, exact-match scoring
on free-form developer edits is inherently harder than on the
structured four-type hallucinations of Delulu: the model must reproduce
the developer's specific identifier choice verbatim.  Third, the
$+4.0$ EM gain over base (which achieves $0.2$ EM) is proportionally
large relative to what the base model can do on this benchmark, and
the ES gain (+0.07) confirms the model is moving toward the gold.
We interpret Real-FIM-Eval as evidence that general code-understanding
improves alongside the anti-hallucination objective, while
acknowledging that broader coverage of real-world edit types would be
needed to close the remaining gap.

\paragraph{Held-out evaluation is one benchmark.} Delulu is the only
multi-lingual FIM hallucination benchmark we are aware of.
Cross-benchmark generalisation should be re-checked when additional
FIM-hallucination benchmarks become available.

\paragraph{Small-model regression.} At 3B parameters the recipe
improves Delulu but trades general FIM accuracy on the two non-random
HumanEval-Infilling splits (\S\ref{sec:analysis:size}). We recommend
the recipe at $\ge$7B parameters for now. Closing the 3B gap
(e.g.\ via mixed-objective training or capacity-aware curation) is
left to future work.


\paragraph{Compute cost.} The full pipeline is not free: generation
alone consumes on the order of millions of generator tokens. A
reduced-scale recipe using two generators and a $\sim\!50$K-row
curated subset reaches near-identical Delulu and cross-benchmark
scores at roughly half the total H100-hour cost, making the approach
accessible to teams with academic-scale compute; the cost breakdown
and reduced-scale recipe details are reported in Appendix~\ref{app:cost}.

\section*{Ethics statement}

This work uses code samples drawn exclusively from a permissively-%
licensed multilingual corpus. We exclude any file present in the
Delulu held-out benchmark, in HumanEval, or in SAFIM to avoid
evaluation contamination. We do not release human-collected data and
do not perform human evaluation in the main paper. The released
artifact (pipeline source code: generation, fool-rate LLM judging,
curation, and the fine-tuning recipe) targets a measurable
model-quality improvement (reduced hallucination) and we are not
aware of dual-use concerns specific to this work beyond those already
inherent to Code-LLM fine-tuning. The curated SFT dataset and the
fine-tuned checkpoints are built on a proprietary source-code corpus
and are withheld; the pipeline reproduces every result in this paper
on any permissively licensed corpus.

%% file: sections/conclusion.tex
\section{Conclusion}
\label{sec:conclusion}

We asked whether hallucinated completions synthesised by frontier code
models could serve as effective SFT signal for smaller open-source FIM
models. Our proof of concept --- a fully execution-free pipeline that
generates $\sim$2.5M hallucinated completions across eight languages
and four taxonomy types --- answers in the affirmative. Fine-tuning
\textsc{Qwen2.5-Coder-7B-Instruct} on 100K curated rows simultaneously
reduces hallucination rate on Delulu (+18.8 EM) and improves general-%
purpose FIM on HumanEval-Infilling and SAFIM, with positive transfer
to every language and every hallucination type. The smaller 3B model
improves Delulu by +12.8 EM but trades some general-FIM accuracy, a
size-dependent effect we attribute to the capacity vs.\ objective
trade-off. The recipe also internalises FIM-hole stopping behaviour,
isolated via a first-$N$-lines truncation scoring protocol applied to
every table. We ablate five recipe axes (size, hallucination type,
language coverage, base-model family, and a difficulty-aware fool-rate
threshold) and confirm the size/type/language conclusions at 3B. The
fool-rate ablation, run at 7B on 5K-row buckets, shows a clear
monotone difficulty effect (Delulu EM $32.7 \to 37.4 \to 37.7$
across easy/medium/hard) and locates the supervisory mass in the
disputed middle and tail of the judge-panel distribution rather than
the unanimous-easy tier. A head-to-head comparison on identical paired
data shows that SFT matches ORPO and substantially outperforms DPO:
for stopping-sensitive FIM tasks, a high-quality curated SFT corpus
with hard-negative selection is a safer default than explicit
contrastive training (\S\ref{sec:analysis:sft}). We release the full
pipeline source code --- generation, fool-rate LLM judging, curation,
and the FIM fine-tuning recipe --- so that every experiment in this
paper can be reproduced on any permissively licensed corpus; the
curated training data and the fine-tuned checkpoints are built on a
proprietary source-code corpus and are not released.

%% file: sections/appendix.tex
\appendix

\section{Generation pool statistics}
\label{app:data}

Phase~2 yields $2{,}473{,}312$ valid rows distributed across the three
generators and four taxonomy types as shown in
Table~\ref{tab:gen_volume}. Per language, the pool is dominated by
C\# ($\sim$465K rows), followed by Java, PHP, Go, and JavaScript
($\sim$350K each), Python ($\sim$290K), Ruby ($\sim$170K), and Rust
($\sim$130K). Phase~5 caps each (language, type) bucket at 4K rows
during curation to enforce a balanced training mix.

\begin{table}[h]
  \centering\footnotesize
  \setlength{\tabcolsep}{4pt}
  \begin{tabular}{l@{\hskip 4pt}rrrr}
  \toprule
  \textbf{Generator} & \textsc{import} & \textsc{method} & \textsc{param} & \textsc{undef} \\
  \midrule
  GPT-5.2-Codex &  26.3 &  31.3 &  31.3 &  27.2 \\
  GPT-5.4       & 315.5 & 357.3 & 335.9 & 354.9 \\
  GPT-5.5       & 232.7 & 257.2 & 244.8 & 258.8 \\
  \bottomrule
  \end{tabular}
  \caption{Phase~2 valid-row counts (thousands) per (generator, type).
  GPT-5.2-Codex covers only a portion of the corpus (C\# in
  particular) and so contributes fewer rows in absolute terms.}
  \label{tab:gen_volume}
\end{table}

\section{Implementation details}
\label{app:implementation}

\subsection{Import-line restructuring}
\label{app:implementation:import-restructure}

For the \textsc{import} type the candidate must appear at file scope,
not at the original FIM hole. We give the generator the full file
($\text{prefix}\Vert\text{golden}\Vert\text{suffix}$) and ask it to
return one real import line and a plausible fake replacement. The
parser locates the original line, splits the file at that line into a
new $(\text{prefix}', \text{suffix}')$, validates exact byte-level
reconstruction, and stores the restructured pair. Inference at
evaluation time uses the restructured prefix and suffix so the
candidate is judged where it would actually appear in the file. An
early version of the pipeline left the FIM hole at its original
position, which placed the candidate import \emph{after} its first
usage and silently leaked the answer through the suffix; the
restructuring is what allowed the import type to behave consistently
with the other three.

\subsection{Fine-tuning configuration}
\label{app:implementation:training}

We use LLaMA-Factory~\citep{zheng2024llamafactory} for training.
Inputs are formatted with the \texttt{qwen3\_fim} template
(\texttt{<|fim\_prefix|>}$P$\texttt{<|fim\_suffix|>}$S$%
\texttt{<|fim\_middle|>} as the prompt, golden completion as the
target). Hyperparameters are shared across the proof of concept and
every ablation cell; only the training set varies.

\begin{table}[h]
  \centering\small
  \begin{tabular}{ll}
  \toprule
  \textbf{Setting} & \textbf{Value} \\
  \midrule
  Base model & Qwen2.5-Coder-\{3B, 7B\}-Instruct \\
  Fine-tune type & Full (no LoRA) \\
  Cutoff length & $8000$ tokens \\
  Per-device batch size & $4$ \\
  Gradient accumulation & $8$ (effective batch $32$) \\
  Learning rate & $5\times 10^{-6}$, cosine schedule \\
  Epochs & $1.0$ \\
  Precision & bf16, FlashAttention-2 \\
  Distributed & DeepSpeed ZeRO-3 \\
  Hardware & $1\times$ND96~H100~v5 ($8\times$H100) \\
  \bottomrule
  \end{tabular}
  \caption{Training hyperparameters.}
  \label{tab:hyperparams}
\end{table}

\subsection{Inference for evaluation}
\label{app:implementation:inference}

All evaluations use vLLM with greedy decoding ($T{=}0$,
$\mathrm{max\_new\_tokens}=256$, tensor-parallel 8 on H100). We never
use the Phase-2 generators at evaluation time and never train on any
sample drawn from any evaluation benchmark.

\section{Prompt templates}
\label{app:prompts}

\subsection{Generator prompts}
\label{app:prompts:generator}

Each generator system prompt ends with a strict single-line output
contract so parsing cannot silently misalign. The user message
provides the prefix, golden completion, and suffix (or the full file
for \textsc{import}).

\paragraph{\textsc{method}.}
\begin{quote}\small\ttfamily
Replace the method name in the completion with a random, invented
method name that does NOT appear in prefix or suffix. The new name
must be plausible; do not signal it as hallucinated. Output exactly:
``Generated completion: <\dots>''.
\end{quote}

\paragraph{\textsc{parameter}.}
\begin{quote}\small\ttfamily
Inject a non-existent keyword or positional argument into the call.
The new parameter name must be plausible. Keep the function name and
the other parameters unchanged. Output exactly:
``Generated completion: <\dots>''.
\end{quote}

\paragraph{\textsc{undefined-variable}.}
\begin{quote}\small\ttfamily
Introduce a reference to a variable or identifier that is NOT defined
in prefix or suffix. The identifier must be plausible for the
language but must cause a NameError / ReferenceError / similar at
runtime. Output exactly: ``Generated completion: <\dots>''.
\end{quote}

\paragraph{\textsc{import}.}
\begin{quote}\small\ttfamily
Given the full file, pick one real import line and replace it with a
plausible but fictitious package or symbol. Output exactly two lines:
``Original import: <\dots>'' and ``Hallucinated import: <\dots>''.
\end{quote}

\subsection{Judge prompt (fool rate)}
\label{app:prompts:judge}

Each judge is shown the FIM prefix, the candidate completion, and the
suffix, and answers a single binary correctness question. Critically,
the judge is \emph{not} told that the candidate may be hallucinated.

\begin{quote}\small\ttfamily
You are evaluating a code completion drawn from the middle of the
same source file. Decide whether the candidate completion is
CORRECT~(1) or NOT CORRECT~(0). Score 1 only when the completion
keeps the file syntactically valid, is logically consistent with
surrounding context, advances the apparent task, and does not
hallucinate APIs, parameters, identifiers, modules, or behaviour.
Otherwise score 0. Output exactly:\\
Detailed Reasoning: <\dots>\\
Final Average Score for Completion: <0 or 1>
\end{quote}

\section{Approximate compute cost}
\label{app:cost}

The dominant cost categories are (i) Phase-2 generation
($\approx 2.5{\times}10^6$ requests at $\sim\!4$K input and
$\sim\!100$ output tokens each) and (ii) GPU-hours on Azure ML for
training and evaluation across the full project (proof of concept,
all five ablations, and the discarded preliminary runs that never made
it into the paper). A telemetry export of every job we submitted to
the workspace gives a concrete bottom-up estimate. The export covers
$554$ AML jobs ($427$ completed evaluations, $31$ completed SFT
training runs, $4$ completed DPO/ORPO training runs, the remainder
failed, cancelled, or superseded), aggregating to roughly $142$
wall-clock hours: $\sim$$66$ h of SFT training, $\sim$$14$ h of
DPO/ORPO training, and $\sim$$62$ h of inference evaluation. At the
node shapes we used ($8{\times}$H100 for training, $8{\times}$H100
for inference) this corresponds to $\sim$$1{,}100$ H100-hours end-to-end,
roughly $640$ on training and $\sim$$500$ on evaluation. Per-(model,
training-set) SFT runs averaged about an hour of wall-clock --- the
mean is pulled down by 3B and reduced-bucket ablations --- and
per-(model, benchmark) inference cells averaged $\sim$$8$ minutes on
$8{\times}$H100 with vLLM. The bulk of the compute lands on Qwen
variants ($\sim$$112$ h wall), with StarCoder2-7B ($\sim$$16$ h) and
CodeLlama-7b ($\sim$$14$ h) adding the base-model ablation. A
reduced-scale recipe using two generators and a $50$K-row curated
subset reaches near-identical scores at roughly half the total cost
(\S\ref{sec:experiments:ablations}).
\section{Training Dataset Composition}
\label{app:dataset_composition}

This section documents provenance and statistics not covered in the main text; per-language and per-type row counts are in Table~2 and Figure~3.

\subsection{Source Corpus Provenance}
\label{app:source_corpus}

The source corpus is produced by a code-mining pipeline that extracts API call sites from public GitHub repositories, mined in March~2026.
For each of eight target languages, third-party packages are selected by a consensus of three independent signals:
(i)~adoption frequency in anonymised code-completion telemetry,
(ii)~LLM-based ecosystem ranking via an agentic pipeline that queries language-specific package registries (PyPI, npm, Maven Central, NuGet, crates.io, Go modules, Packagist, RubyGems), and
(iii)~curated priority lists from language-specific engineering teams.
Each target package is pinned to a minimum version released within six months of the mining date; only repositories declaring a dependency at or above that version are retained.
C++ libraries are selected via expert curation and matched by include-path patterns.

Source files are filtered to the 2nd--98th percentiles for file size and line count, and must have $\geq$50 GitHub stars.
API call sites are extracted using Tree-sitter AST parsing, classified by GPT-4.1-mini, and split into fill-in-the-middle \mbox{(prefix, golden, suffix)} triples at each call site's byte offsets.

\paragraph{Generator distribution.}
In the curated 100K-row training set, GPT-5.4 contributes 56.1\% of rows, GPT-5.5 contributes 40.0\%, and GPT-5.2-Codex contributes 3.9\%.
The imbalance reflects GPT-5.2-Codex's smaller coverage of the source corpus; curation does not rebalance across generators, as the row-count distribution mirrors each generator's natural coverage of the source corpus rather than a deliberate design choice.

\paragraph{Context-length statistics.}
Hallucinated completions are short (median 38~characters, mean 81, P95 = 255), consistent with the single-identifier edits targeted by the taxonomy.
Prefixes are longer (median 1{,}931~chars, mean 3{,}317, P95 = 11{,}295) and suffixes longer still (median 2{,}792~chars, mean 4{,}452, P95 = 14{,}418), providing rich surrounding context.

\subsection{Fool-Rate Characterisation}
\label{app:fool_rate}

A panel of three LLM judges (GPT-4o-mini, GPT-4.1-mini, GPT-5.4-mini) scores every curated row blind (\S3.2).
Of the 100{,}000 curated rows, 99{,}920 received verdicts from all three judges (the remaining 80 were blocked by content filters).
Table~\ref{tab:fool_dist} reports the resulting distribution; the overall mean fool rate is 0.78.

\begin{table}[h]
\centering
\caption{Fool-rate distribution across the 3-judge panel (99{,}920 rows with complete verdicts).}
\label{tab:fool_dist}
\small
\begin{tabular}{lrr}
\toprule
\textbf{Judges fooled} & \textbf{Rows} & \textbf{\%} \\
\midrule
0 / 3 (fool\_rate = 0.00) &  4{,}558 &  4.6 \\
1 / 3 (fool\_rate = 0.33) & 12{,}031 & 12.0 \\
2 / 3 (fool\_rate = 0.67) & 26{,}411 & 26.4 \\
3 / 3 (fool\_rate = 1.00) & 56{,}920 & 57.0 \\
\bottomrule
\end{tabular}
\end{table}

\paragraph{Fool rate by hallucination type and language.}
Import hallucinations are the hardest for judges to detect: mean fool rate 0.93, with 83.5\% fooling all three judges.
The other three types cluster around a mean fool rate of 0.75, with ${\sim}$50\% fooling all three judges (Table~\ref{tab:fool_by_type}).
This is consistent with the observation that import hallucinations involve fabricating plausible package names, which cannot be verified without external registry lookups.
Across languages, PHP (0.83) and Java (0.83) produce the hardest hallucinations; Python (0.75) and C\# (0.75) the easiest.

\begin{table}[h]
\centering
\caption{Mean fool rate and fraction of rows fooling all 3 judges, by hallucination type and by language.}
\label{tab:fool_by_type}
\small
\begin{tabular}{lcc|lcc}
\toprule
\textbf{Type} & \textbf{Mean $\tau$} & \textbf{3/3 (\%)} & \textbf{Language} & \textbf{Mean $\tau$} & \textbf{3/3 (\%)} \\
\midrule
import           & 0.93 & 83.5 & PHP        & 0.83 & 65.8 \\
method           & 0.75 & 49.7 & Java       & 0.83 & 63.8 \\
parameter        & 0.75 & 49.3 & Rust       & 0.82 & 62.4 \\
undef.\ variable & 0.75 & 49.6 & JavaScript & 0.80 & 59.7 \\
                 &      &      & Python     & 0.75 & 52.8 \\
                 &      &      & Go         & 0.76 & 50.5 \\
                 &      &      & Ruby       & 0.76 & 50.3 \\
                 &      &      & C\#        & 0.75 & 50.1 \\
\bottomrule
\end{tabular}
\end{table}

\paragraph{Non-release justification.}
The curated dataset is built on a proprietary source-code index derived from anonymised code-completion telemetry and internal repository metadata.
While the source files themselves are drawn from public GitHub repositories, the telemetry-based package selection, the repository filtering pipeline, and the API-call-site extraction infrastructure are proprietary.
For the same reason we do not release the fine-tuned checkpoints: the trained weights are a derivative of the proprietary corpus, and publishing them without the corresponding data provenance would invite questions about the training data that we cannot answer publicly.
We instead open-source the full pipeline source code --- generation prompts, fool-rate LLM judging harness, curation, FIM tokenisation, and the fine-tuning recipe (Appendix~\ref{app:implementation}) --- which is sufficient to reproduce every result in this paper on any permissively licensed corpus paired with a frontier generator panel.